\documentclass{article}

\usepackage[final]{nips_2016}

\usepackage[utf8]{inputenc} 
\usepackage[T1]{fontenc}    
\usepackage{url}            
\usepackage{booktabs}       
\usepackage{amsfonts}       
\usepackage{nicefrac}       
\usepackage{microtype}      

\usepackage{natbib} 
\usepackage{graphicx} 
\graphicspath{{images/}{gm_bnmtf/}{greedy_search/}}
\usepackage{amssymb} 
\usepackage{amsmath} 
\usepackage{caption} 
\usepackage{subcaption} 
\usepackage{booktabs} 

\newcommand{\R}{\boldsymbol R}
\newcommand{\E}{\boldsymbol E}
\newcommand{\U}{\boldsymbol U}
\newcommand{\V}{\boldsymbol V}
\newcommand{\F}{\boldsymbol F}
\renewcommand{\S}{\boldsymbol S}
\newcommand{\G}{\boldsymbol G}

\newcommand{\btheta}{\boldsymbol \theta}

\newcommand{\lambdaUik}{\lambda_{ik}^U}
\newcommand{\lambdaVjk}{\lambda_{jk}^V}
\newcommand{\lambdaFik}{\lambda_{ik}^F}
\newcommand{\lambdaSkl}{\lambda_{kl}^S}
\newcommand{\lambdaGjl}{\lambda_{jl}^G}

\newcommand{\muUik}{\mu_{ik}^U}
\newcommand{\muVjk}{\mu_{jk}^V}

\newcommand{\tauUik}{\tau_{ik}^U}
\newcommand{\tauVjk}{\tau_{jk}^V}

\newcommand{\expFik}{\widetilde{F_{ik}}}
\newcommand{\expSkl}{\widetilde{S_{kl}}}
\newcommand{\expGjl}{\widetilde{G_{jl}}}

\newcommand{\sumk}{\sum_{k=1}^K}
\newcommand{\sumexclk}{\sum_{k' \neq k}}
\newcommand{\suml}{\sum_{l=1}^L}
\newcommand{\sumexcll}{\sum_{l' \neq l}}

\newcommand{\expdiffTRI}{\mathbb{E}_q \left[ ( R_{ij} - \F_i \cdot \S \cdot \G_j )^2 \right] }

\newcommand{\diffexpTRI}{ \left( R_{ij} - \sumk \suml \expFik \expSkl \expGjl \right) }

\title{Fast Bayesian Non-Negative Matrix Factorisation and Tri-Factorisation}

\author{
  Thomas Brouwer \\
  Computer Laboratory \\
  University of Cambridge \\
  United Kingdom \\
  \texttt{tab43@cam.ac.uk} \\
  \And
  Jes Frellsen \\
  Department of Engineering \\
  University of Cambridge \\
  United Kingdom \\
  \texttt{jf519@cam.ac.uk} \\
  \And
  Pietro Lio' \\
  Computer Laboratory \\
  University of Cambridge \\
  United Kingdom \\
  \texttt{pl219@cam.ac.uk} \\
}

\begin{document}

\maketitle

\begin{abstract}
	We present a fast variational Bayesian algorithm for performing non-negative matrix factorisation and tri-factorisation.
 	We show that our approach achieves faster convergence per iteration and timestep (wall-clock) than Gibbs sampling and non-probabilistic approaches, and do not require additional samples to estimate the posterior.
 	We show that in particular for matrix tri-factorisation convergence is difficult, but our variational Bayesian approach offers a fast solution, allowing the tri-factorisation approach to be used more effectively.
\end{abstract}


\section{Introduction}
Non-negative matrix factorisation methods \cite{Lee1999} have been used extensively in recent years to decompose matrices into latent factors, helping us reveal hidden structure and predict missing values. In particular we decompose a given matrix into two smaller matrices so that their product approximates the original one. The non-negativity constraint makes the resulting matrices easier to interpret, and is often inherent to the problem -- such as in image processing or bioinformatics (\cite{Lee1999,Wang2013}). Some approaches approximate a maximum likelihood (ML) or maximum a posteriori (MAP) solution that minimises the difference between the observed matrix and the decomposition of this matrix. This gives a single point estimate, which can lead to overfitting more easily and neglects uncertainty. Instead, we may wish to find a full distribution over the matrices using a Bayesian approach, where we define prior distributions over the matrices and then compute their posterior after observing the actual data.

\citet{Schmidt2009} presented a Bayesian model for non-negative matrix factorisation that uses Gibbs sampling to obtain draws of these posteriors, with exponential priors to enforce non-negativity. Markov chain Monte Carlo (MCMC) methods like Gibbs sampling rely on a sampling procedure to eventually converge to draws of the desired distribution -- in this case the posterior of the matrices. This means that we need to inspect the values of the draws to determine when our method has converged (burn-in), and then take additional draws to estimate the posteriors. 

We present a variational Bayesian approach to non-negative matrix factorisation, where instead of relying on random draws we obtain a deterministic convergence to a solution. We do this by introducing a new distribution that is easier to compute, and optimise it to be as similar to the true posterior as possible. 
We show that our approach gives faster convergence rates per iteration and timestep (wall-clock) than current methods, and is less prone to overfitting than the popular non-probabilistic approach of \citet{Lee2000}.

We also consider the problem of non-negative matrix tri-factorisation, first introduced by \citet{Ding2006}, where we decompose the observed dataset into three smaller matrices, which again are constrained to be non-negative. Matrix tri-factorisation has been explored extensively in recent years, for example for collaborative filtering (\cite{Chen2009}) and clustering genes and phenotypes (\cite{Hwang2012}). 
We present a fully Bayesian model for non-negative matrix tri-factorisation, extending the matrix factorisation model to obtain both a Gibbs sampler and a variational Bayesian algorithm for inference. We show that convergence is even harder, and that the variational approach provides significant speedups (roughly four times faster) compared to Gibbs sampling.


\section{Non-Negative Matrix Factorisation}
We follow the notation used by \citet{Schmidt2009} for non-negative matrix factorisation (NMF), which can be formulated as decomposing a matrix $ \R \in \mathbb{R}^{I \times J} $ into two latent (unobserved) matrices $ \U \in \mathbb{R}_+^{I \times K} $ and $ \V \in \mathbb{R}_+^{J \times K} $. In other words, solving $ \R = \U \V^T + \E $, where noise is captured by matrix $ \E \in \mathbb{R}^{I \times J} $. The dataset $ \R $ need not be complete -- the indices of observed entries can be represented by the set $ \Omega = \left\{ (i,j) \text{ $ \vert $ $ R_{ij} $ is observed} \right\} $. These entries can then be predicted by $\U \V^T$.

We take a probabilistic approach to this problem. We express a likelihood function for the observed data, and treat the latent matrices as random variables. As the likelihood we assume each value of $ \R $ comes from the product of $ \U $ and $ \V $, with some Gaussian noise added,
\begin{equation*}
	R_{ij} \sim \mathcal{N} (R_{ij} | \boldsymbol U_i \cdot \boldsymbol V_j, \tau^{-1} )
\end{equation*}
\noindent where $ \boldsymbol U_i, \boldsymbol V_j $ denote the $i$th and $j$th rows of $\U$ and $\V $, and $ \mathcal{N} (x|\mu,\tau) $
is the density of the Gaussian distribution, with precision $ \tau $. The full set of parameters for our model is denoted $ \btheta = \left\{ \U, \V, \tau \right\} $.

In the Bayesian approach to inference, we want to find the distributions over the parameters $ \btheta $ after observing the data $ D = \lbrace R_{ij} \rbrace_{i,j \in \Omega} $. We can use Bayes' theorem for this, $ p(\btheta|D) \propto p(D|\btheta) p(\btheta) $.
We need priors over the parameters, allowing us to express beliefs for their values -- such as constraining $ \U, \V $ to be non-negative. We can normally not compute the posterior $ p(\btheta|D) $ exactly, but some choices of priors allow us to obtain a good approximation. Schmidt et al. choose an exponential prior over $ \U $ and $ \V $, so that each element in $ U $ and $ V $ is assumed to be independently exponentially distributed with rate parameters $ \lambdaUik, \lambdaVjk > 0 $,
\begin{alignat*}{2}
	U_{ik} \sim \mathcal{E} ( U_{ik} | \lambdaUik)		\quad\quad		V_{jk} \sim \mathcal{E} ( V_{jk} | \lambdaVjk)
\end{alignat*}
\noindent where $ \mathcal{E} ( x | \lambda ) $
is the density of the exponential distribution.
For the precision $ \tau $ we use a Gamma distribution with shape $ \alpha > 0 $ and rate $ \beta > 0 $.

\citet{Schmidt2009} introduced a Gibbs sampling algorithm for approximating the posterior distribution, which relies on sampling new values for each random variable in turn from the conditional posterior distribution. Details on this method can be found in the supplementary materials (Section 1.1).

\paragraph{Variational Bayes for NMF}
Like Gibbs sampling, variational Bayesian inference (VB) is a way to approximate the true posterior $ p(\btheta|D) $. The idea behind VB is to introduce an approximation $q(\btheta)$ to the true posterior that is easier to compute, and to make our variational distribution $q(\btheta)$ as similar to $ p(\btheta|D) $ as possible (as measured by the KL-divergence). We assume the variational distribution $ q(\btheta) $ factorises completely, so all variables are independent, $ q(\btheta) = \prod_{\theta_i \in \btheta} q(\theta_i) $.
%
%
This is called the mean-field assumption. We assume the same forms of $ q(\theta_i) $ as used in Gibbs sampling,
\begin{align*}
	q(\tau) = \mathcal{G} (\tau | \alpha^*, \beta^* )  		\quad\quad\quad		q(U_{ik}) = \mathcal{TN} ( U_{ik} | \muUik, \tauUik ) 		\quad\quad\quad		q(V_{jk}) = \mathcal{TN} ( V_{jk} | \muVjk, \tauVjk )
\end{align*}
\cite{J.M.Bernardo} showed that the optimal distribution for the $i$th parameter, $q^*(\theta_i)$, can be expressed as follows (for some constant $C$), allowing us to find the optimal updates for the variational parameters
\begin{equation*}
	\log q^*(\theta_i) = \mathbb{E}_{q(\btheta_{-i})} \left[ \log p(\btheta, D) \right] + C.
\end{equation*}
Note that we take the expectation with respect to the distribution $ q(\btheta_{-i}) $ over the parameters but excluding the $i$th one. This gives rise to an iterative algorithm: for each parameter $\theta_i$ we update its distribution to that of its optimal variational distribution, and then update the expectation and variance with respect to $ q $. This algorithm is guaranteed to maximise the Evidence Lower Bound (ELBO) 
\begin{equation*}
	\mathcal{L} =  \mathbb{E}_{q} \left[ \log p(\btheta, D) - \log q(\btheta) \right],
\end{equation*}
which is equivalent to minimising the KL-divergence. More details and updates for the approximate posterior distribution parameters are given in the supplementary materials (Section 1.2).


\section{Non-Negative Matrix Tri-Factorisation}
The problem of non-negative matrix tri-factorisation (NMTF) can be formulated similarly to that of non-negative matrix factorisation. We now decompose our dataset $ \R \in \mathbb{R}^{I \times J} $ into three matrices $ \F \in \mathbb{R}_+^{I \times K} $, $ \S \in \mathbb{R}_+^{K \times L} $, $ \G \in \mathbb{R}_+^{J \times L} $, so that $ \R = \F \S \G^T + \E $. We again use a Gaussian likelihood and Exponential priors for the latent matrices.
\begin{alignat*}{2}
	&R_{ij} \sim \mathcal{N} (R_{ij} | \F_i \cdot \S \cdot \G_j, \tau^{-1} )	 		&&\tau \sim \mathcal{G} (\tau | \alpha, \beta ) 		\\
	&F_{ik} \sim \mathcal{E} ( F_{ik} | \lambdaFik)		\quad		S_{kl} \sim \mathcal{E}( S_{kl} | \lambdaSkl)		\quad	&&G_{jl} \sim \mathcal{E}( G_{jl} | \lambdaGjl)
\end{alignat*}
A Gibbs sampling algorithm that can be derived similarly to before. Details can be found in the supplementary materials (Section 1.3).

\subsection{Variational Bayes for NMTF}
Our VB algorithm for tri-factorisation follows the same steps as before, but now has an added complexity due to the term $ \expdiffTRI $. Before, all covariance terms for $ k' \neq k $ were zero due to the factorisation in $ q $, but we now obtain some additional non-zero covariance terms. This leads to the more complicated variational updates given in the supplementary materials (Section 1.4).
\begin{align*}
	\expdiffTRI = \diffexpTRI^2 + \sumk \suml \mathrm{Var}_q \left[ F_{ik} S_{kl} G_{jl} \right] \quad\quad\quad \\
	\quad\quad\quad + \sumk \suml \sumexclk \mathrm{Cov} \left[ F_{ik} S_{kl} G_{jl}, F_{ik'} S_{k'l} G_{jl} \right] + \sumk \suml \sumexcll \mathrm{Cov} \left[ F_{ik} S_{kl} G_{jl}, F_{ik} S_{kl'} G_{jl'} \right]
\end{align*}


\section{Experiments}
To demonstrate the performances of our proposed methods, we ran several experiments on a toy dataset, as well as several drug sensitivity datasets.
For the toy dataset we generated the latent matrices using unit mean exponential distributions, and adding zero mean unit variance Gaussian noise to the resulting product. For the matrix factorisation model we use $ I = 100, J = 80, K = 10 $, and for the matrix tri-factorisation $ I = 100, J = 80, K = 5, L = 5 $.

We also consider a drug sensitivity dataset, which detail the effectiveness of different drugs on cell lines for cancer and tissue types. The Genomics of Drug Sensitivity in Cancer (GDSC v5.0, \cite{Yang2013}) dataset contains 138 drugs and 622 cell lines, with 81\% of entries observed. 

We compare our methods against classic algorithms for matrix factorisation and tri-factorisation. Aside from the Gibbs sampler (G-NMF, G-NMTF) and VB algorithms (VB-NMF, VB-NMTF), we consider the non-probabilistic matrix factorisation (NP-NMF) and tri-factorisation (NP-NMTF) methods introduced by \citet{Lee2000} and \citet{Yoo2009}, respectively. \citet{Schmidt2009} also proposed an Iterated Conditional Modes (ICM-NMF) algorithm for computing an MAP solution, where instead of using draws from the posteriors as updates we set their values to the mode. We also extended this method for matrix tri-factorisation (ICM-NMTF).

\subsection{Convergence speed}
\begin{figure*}[t]
	\begin{subfigure}[t]{1\columnwidth}
		\hspace{100pt}
		\includegraphics[width=0.5\columnwidth]{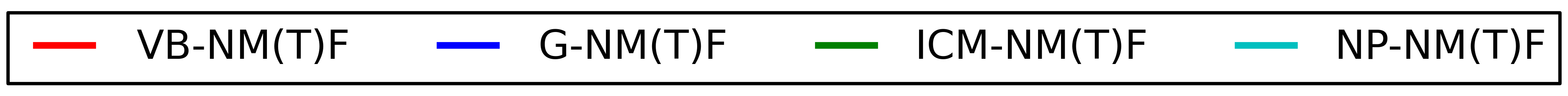}
	\end{subfigure}
	\\
	\begin{subfigure}[t]{0.245 \columnwidth}
		\includegraphics[width=\columnwidth]{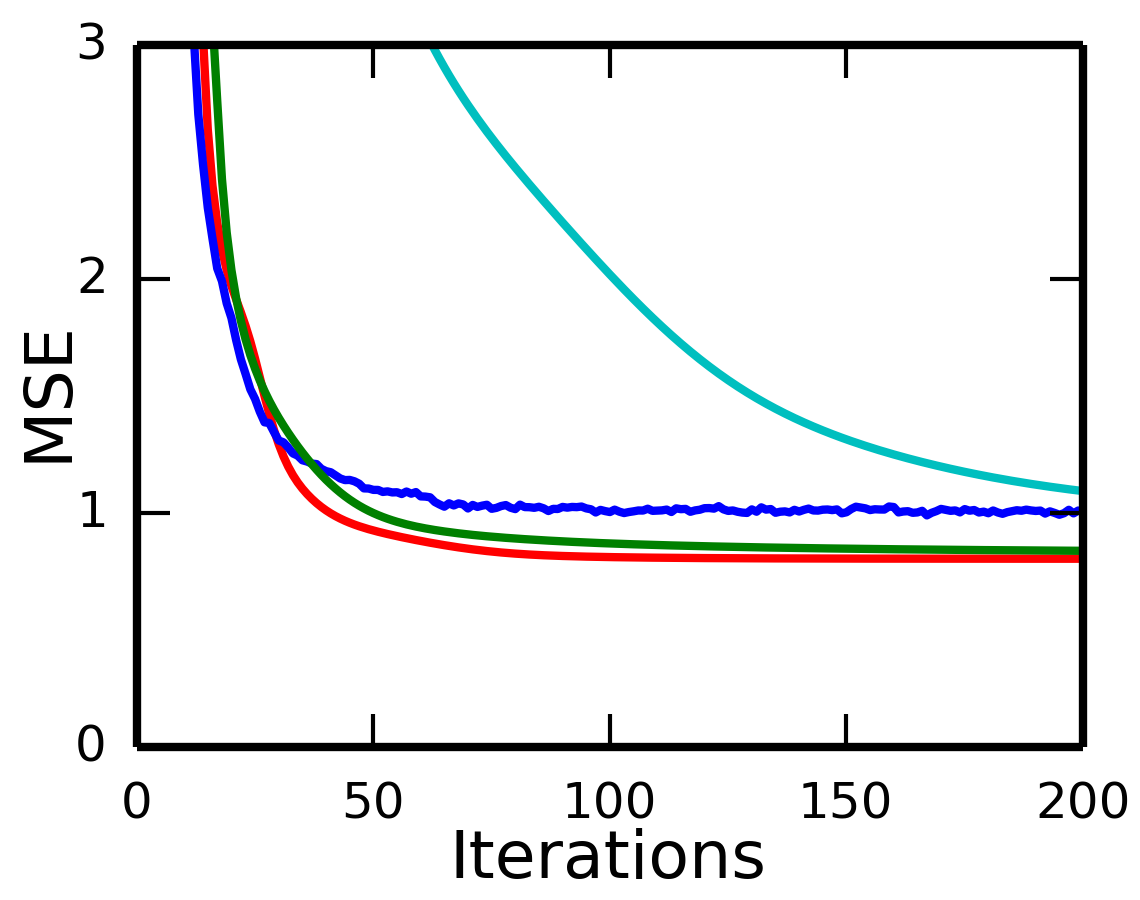}
		\captionsetup{width=0.95\columnwidth}
		\caption{Toy NMF} 
		\label{mse_nmf_convergences}
	\end{subfigure}
	\begin{subfigure}[t]{0.245 \columnwidth}
		\includegraphics[width=\columnwidth]{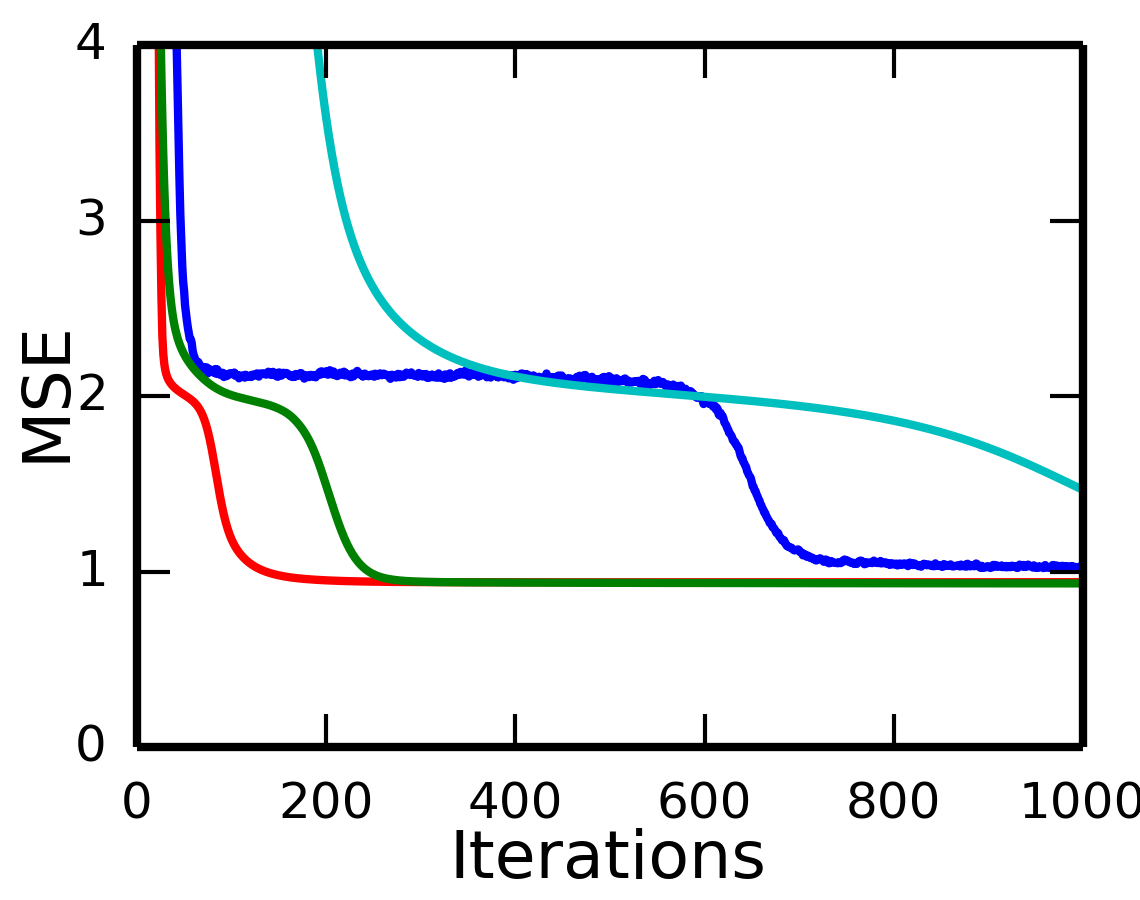}
		\captionsetup{width=0.95\columnwidth}
		\caption{Toy NMTF} 
		\label{mse_nmtf_convergences}
	\end{subfigure}
	\begin{subfigure}[t]{0.245 \columnwidth}
		\includegraphics[width=\columnwidth]{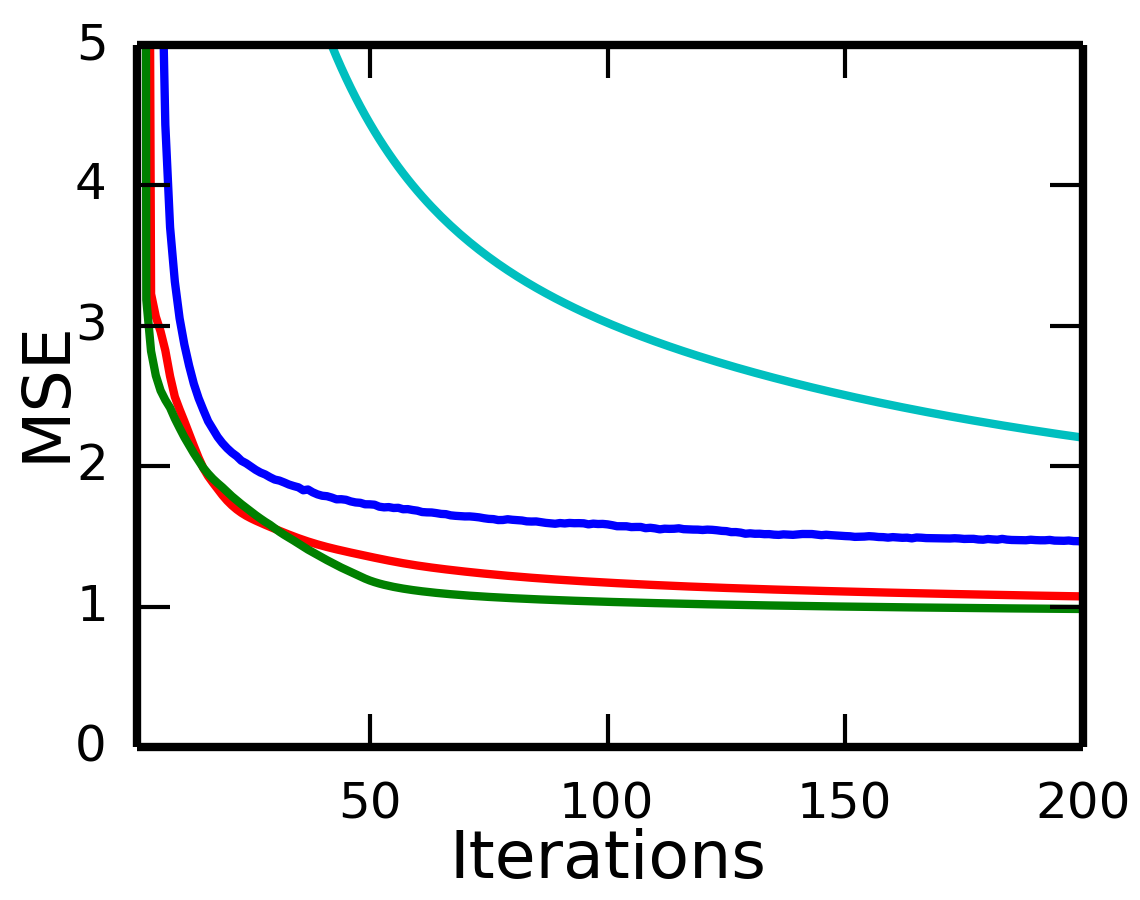}
		\captionsetup{width=0.95\columnwidth}
		\caption{Drug sensitivity NMF} 
		\label{mse_Sanger_nmf_convergences}
	\end{subfigure}
	\begin{subfigure}[t]{0.245 \columnwidth}
		\includegraphics[width=\columnwidth]{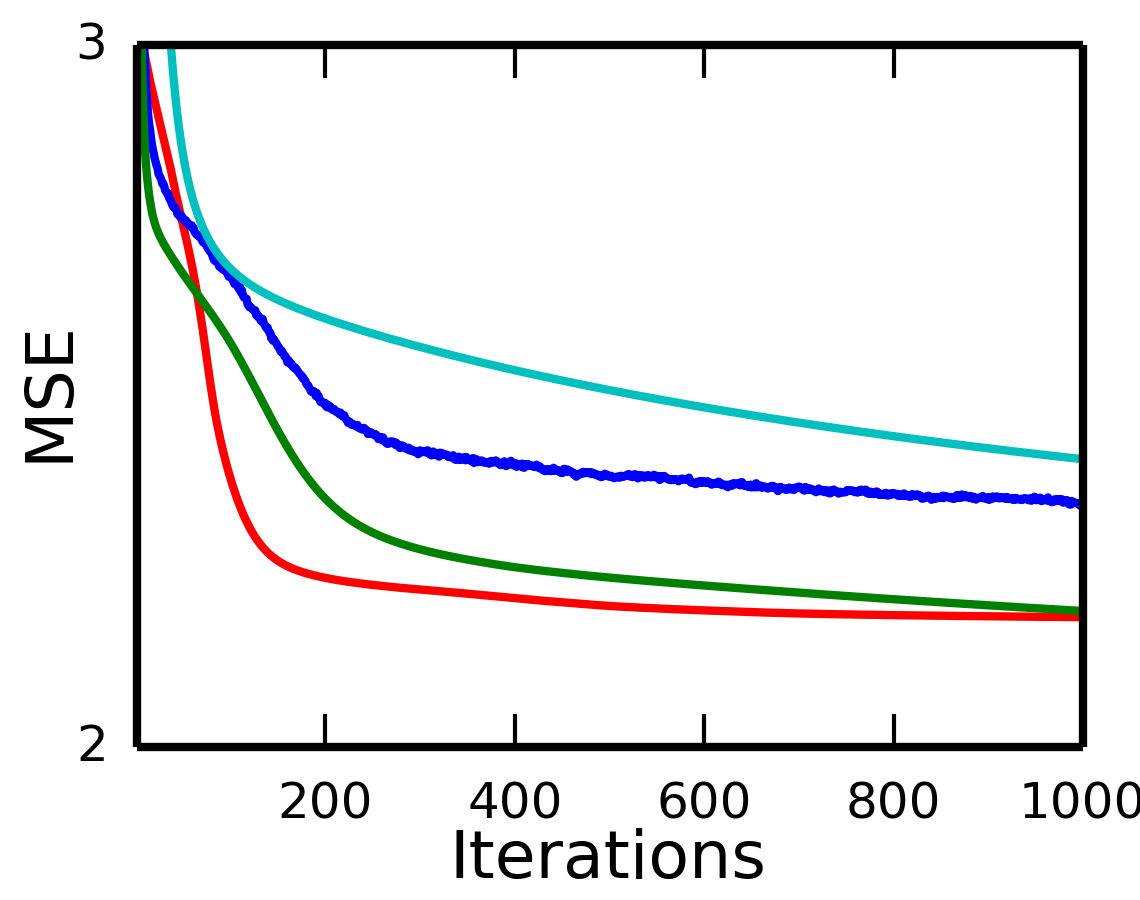}
		\captionsetup{width=0.97\columnwidth}
		\caption{Drug sensitivity NMTF} 
		\label{mse_Sanger_nmtf_convergences}
	\end{subfigure}
	\begin{subfigure}[t]{0.245 \columnwidth}
		\includegraphics[width=\columnwidth]{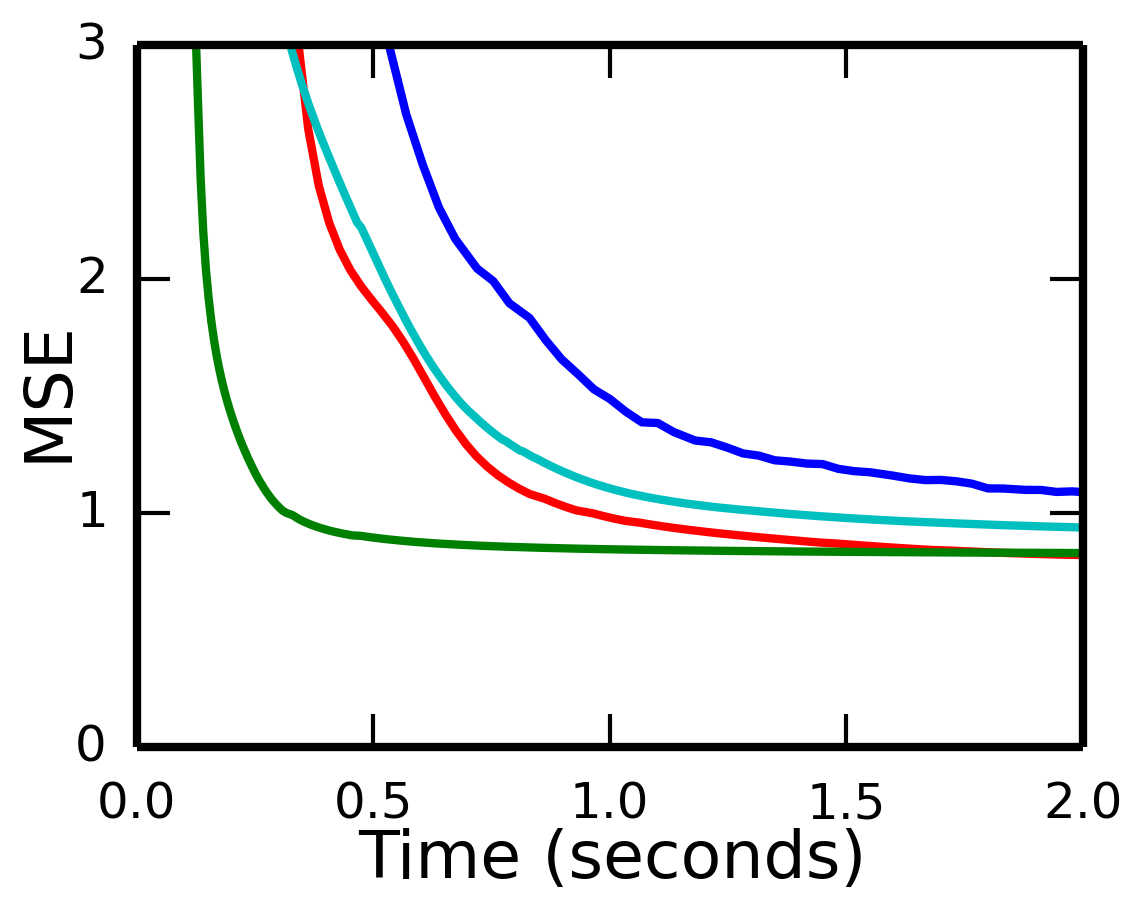}
		\captionsetup{width=0.95\columnwidth}
		\caption{Toy NMF} 
		\label{mse_nmf_times}
	\end{subfigure}
	\begin{subfigure}[t]{0.245 \columnwidth}
		\includegraphics[width=\columnwidth]{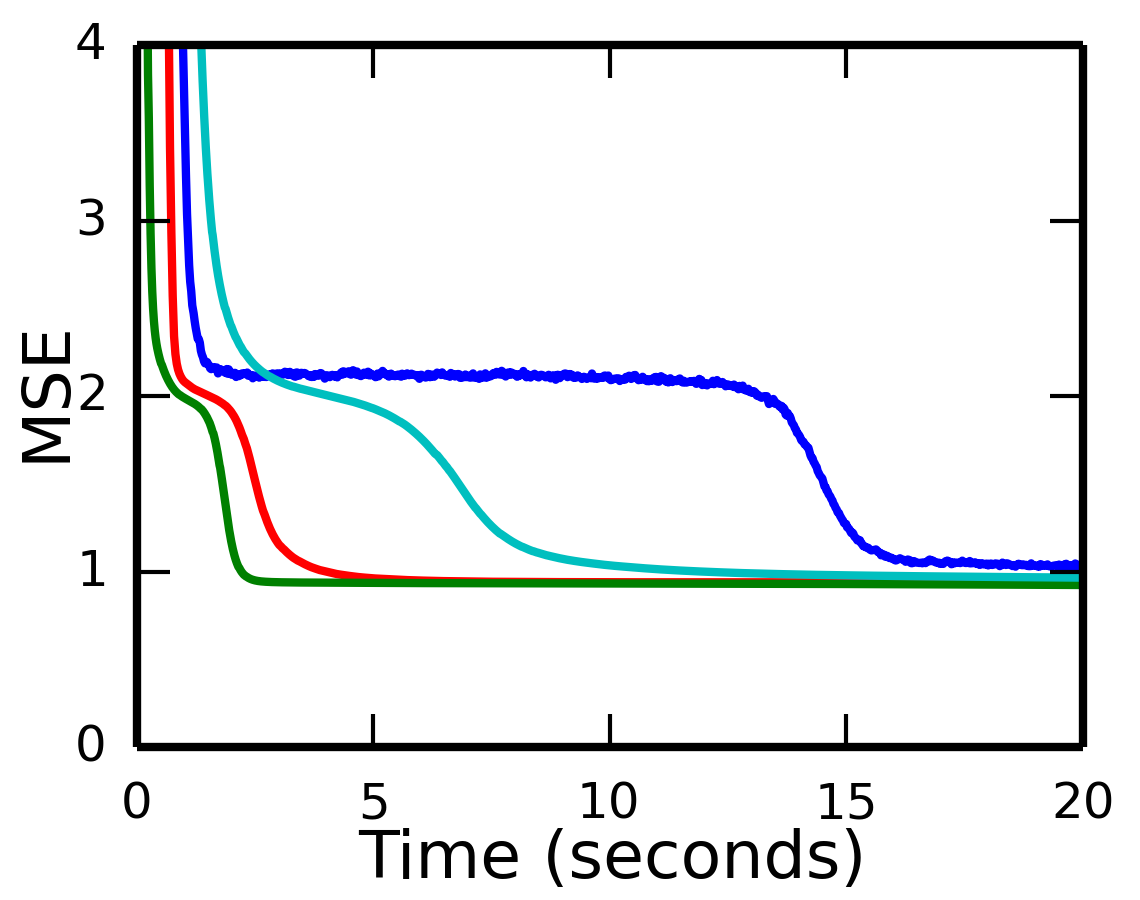}
		\captionsetup{width=0.95\columnwidth}
		\caption{Toy NMTF} 
		\label{mse_nmtf_times}
	\end{subfigure}
	\begin{subfigure}[t]{0.245 \columnwidth}
		\includegraphics[width=\columnwidth]{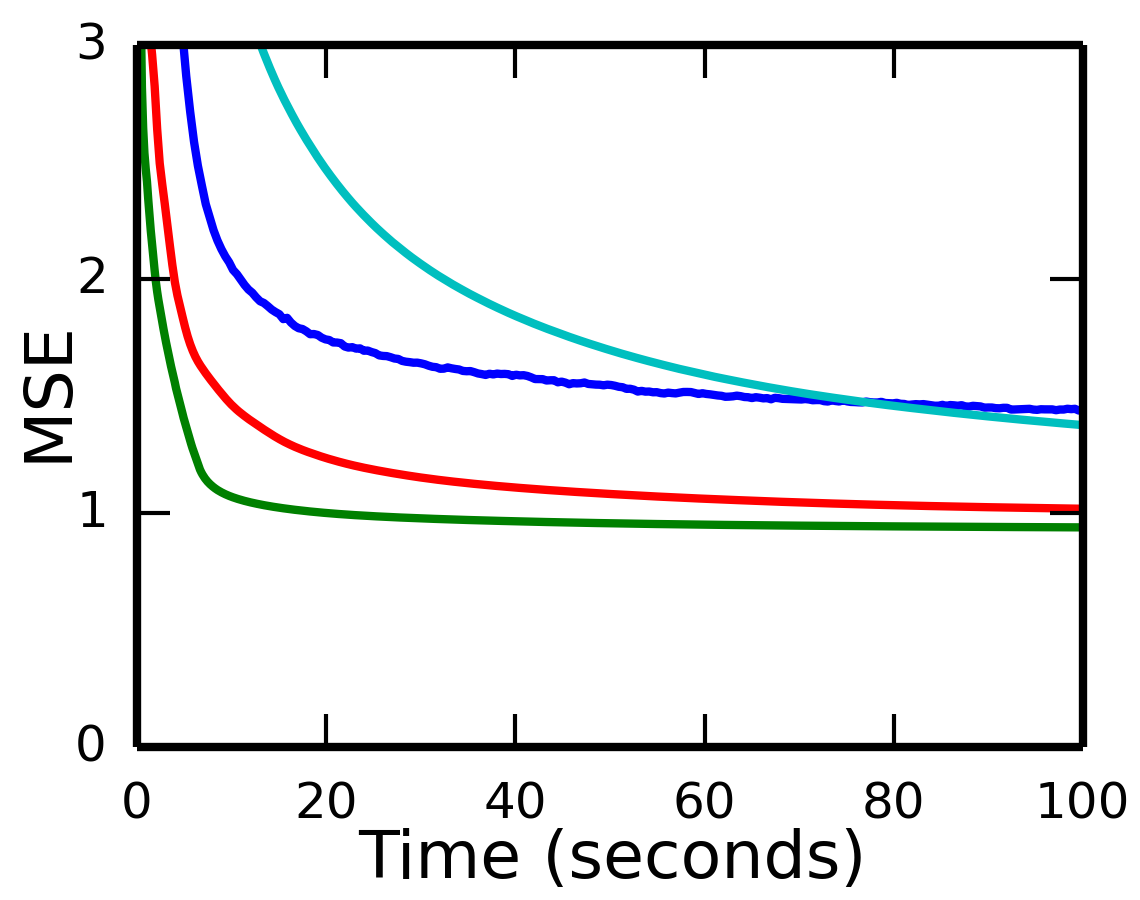}
		\captionsetup{width=0.95\columnwidth}
		\caption{Drug sensitivity NMF} 
		\label{mse_Sanger_nmf_times}
	\end{subfigure}
	\begin{subfigure}[t]{0.245 \columnwidth}
		\includegraphics[width=\columnwidth]{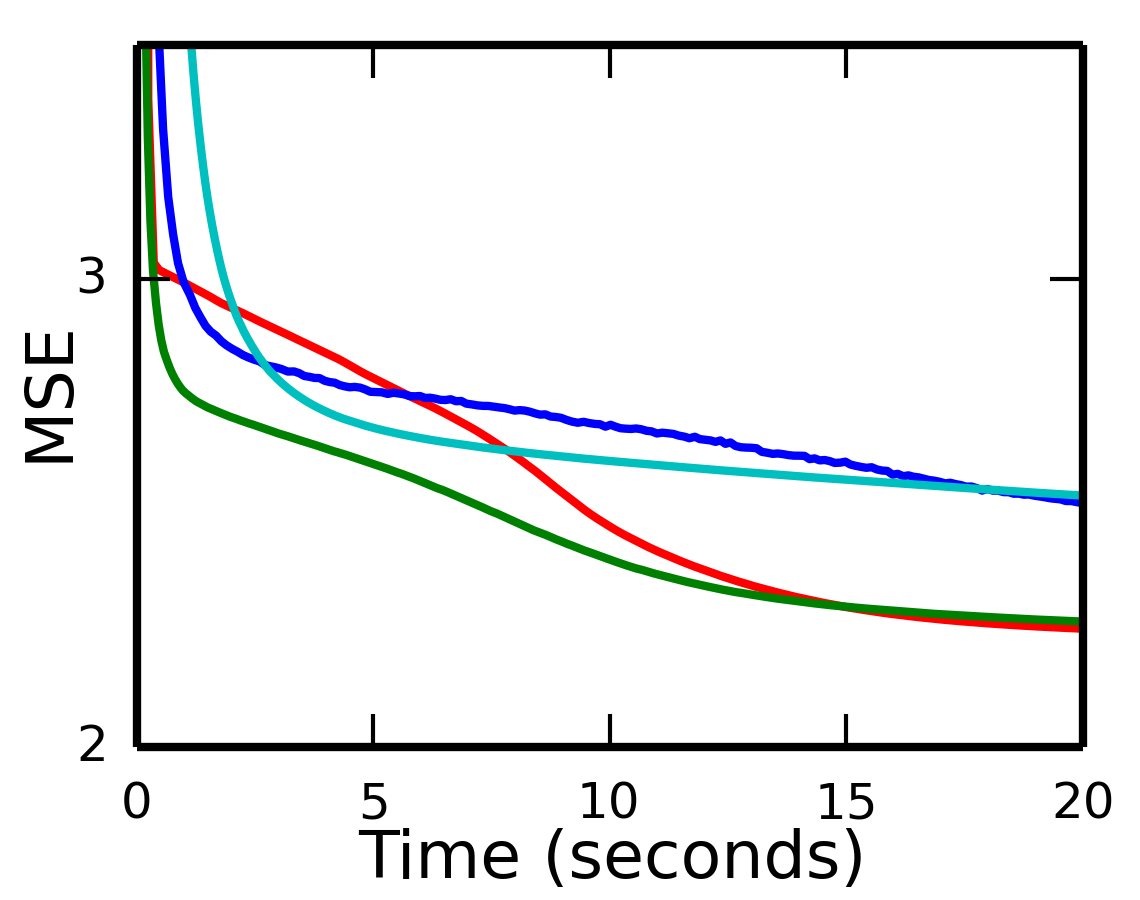}
		\captionsetup{width=0.97\columnwidth}
		\caption{Drug sensitivity NMTF} 
		\label{mse_Sanger_nmtf_times}
	\end{subfigure}
	\captionsetup{width=1\columnwidth}
	\caption{Convergence of algorithms on the toy and GDSC drug sensitivity datasets, measuring the training data fit (mean square error) across iterations (top row) and time (bottom row).}
\end{figure*}

We tested the convergence speed of the methods on both the toy data, giving the correct values for $ K,L $, and on the GDSC drug sensitivity dataset. We track the convergence rate of the error (mean square error) on the training data against the number of iterations taken. This can be found for the toy data in Figures \ref{mse_nmf_convergences} (MF) and \ref{mse_nmtf_convergences} (MTF), and for the GDSC drug sensitivity dataset in Figures \ref{mse_Sanger_nmf_convergences} and \ref{mse_Sanger_nmtf_convergences}. Below that (Figures \ref{mse_nmf_times}-\ref{mse_Sanger_nmtf_times}) is the convergence in terms of time (wall-clock), timing each run 10 times and taking the average (using the same random seed).

We see that our VB methods takes the fewest iterations to converge to the best solution. This is especially the case in the tri-factorisation case, where the best solution is much harder to find (note that all methods initially find a worse solution and get stuck on that for a while), and our variational approach converges seven times faster in terms of iterations taken. We note that time wise, the ICM algorithms can be implemented more efficiently than the fully Bayesian approaches, but returns a MAP solution rather than the full posterior. Our VB method still converges four times faster than the other fully Bayesian approach, and twice as fast as the non-probabilistic method.

\begin{figure*}[t]
	\captionsetup{width=1\columnwidth}
	\begin{subfigure}{0.245 \columnwidth}
		\includegraphics[width=\columnwidth]{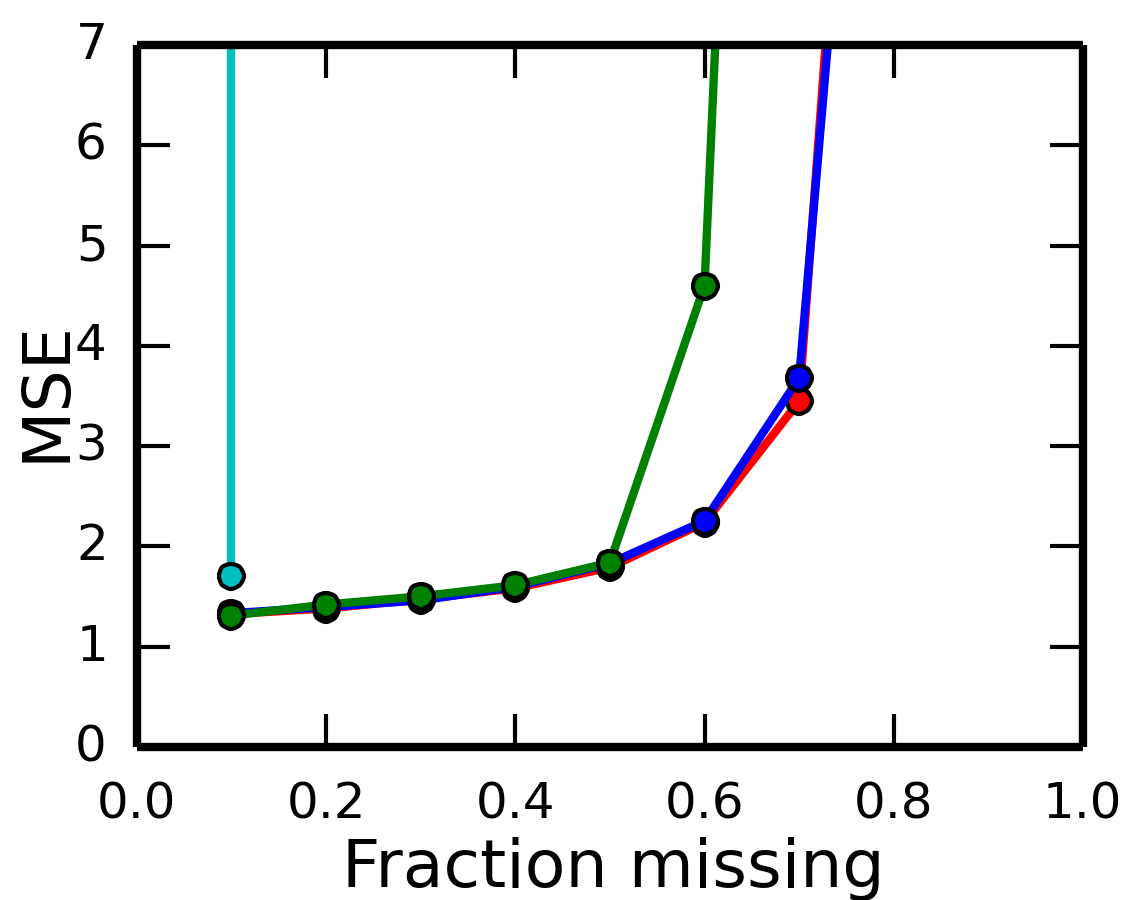}
		\captionsetup{width=0.9\columnwidth}
		\caption{NMF} 
		\label{mse_nmf_missing_values_predictions}
	\end{subfigure} %
	\begin{subfigure}{0.245 \columnwidth}
		\includegraphics[width=\columnwidth]{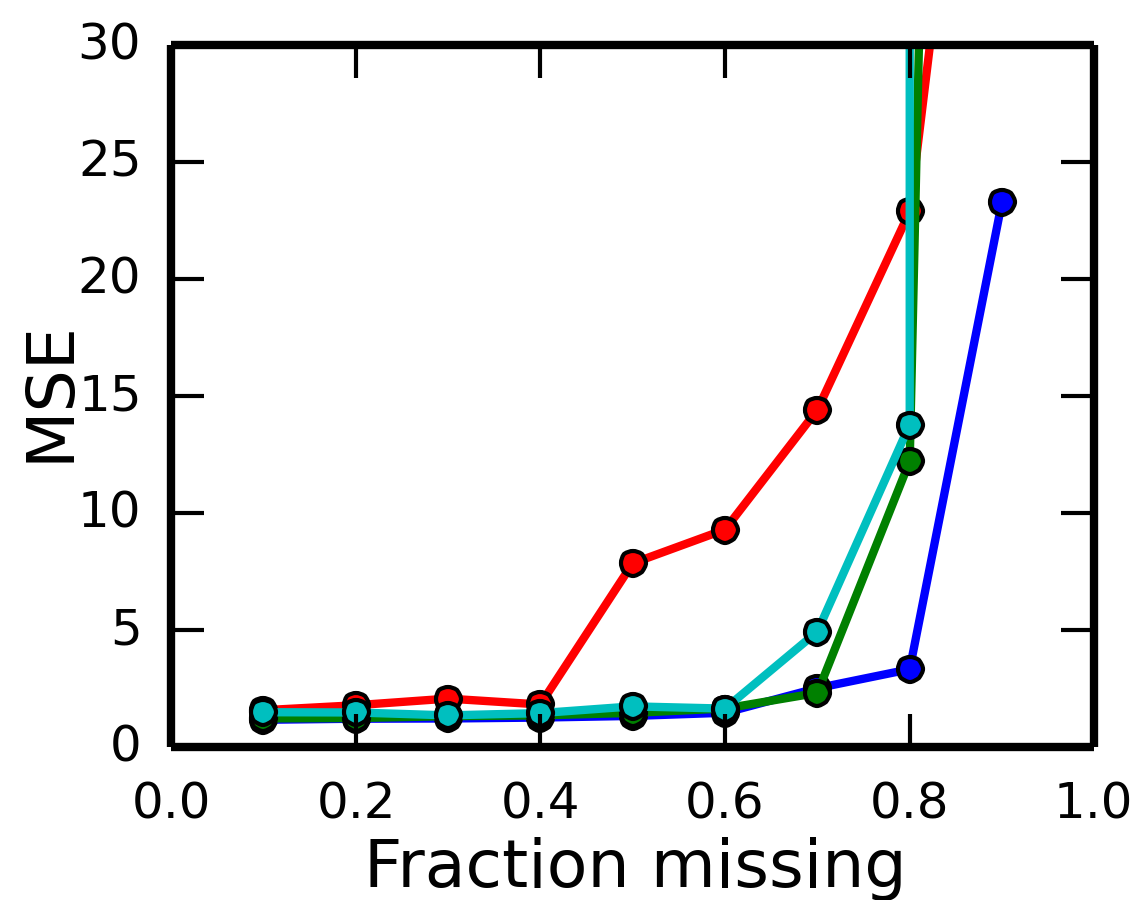}
		\captionsetup{width=0.9\columnwidth}
		\caption{NMTF} 
		\label{mse_nmtf_missing_values_predictions}
	\end{subfigure}
	\begin{subfigure}{0.245 \columnwidth}
		\includegraphics[width=\columnwidth]{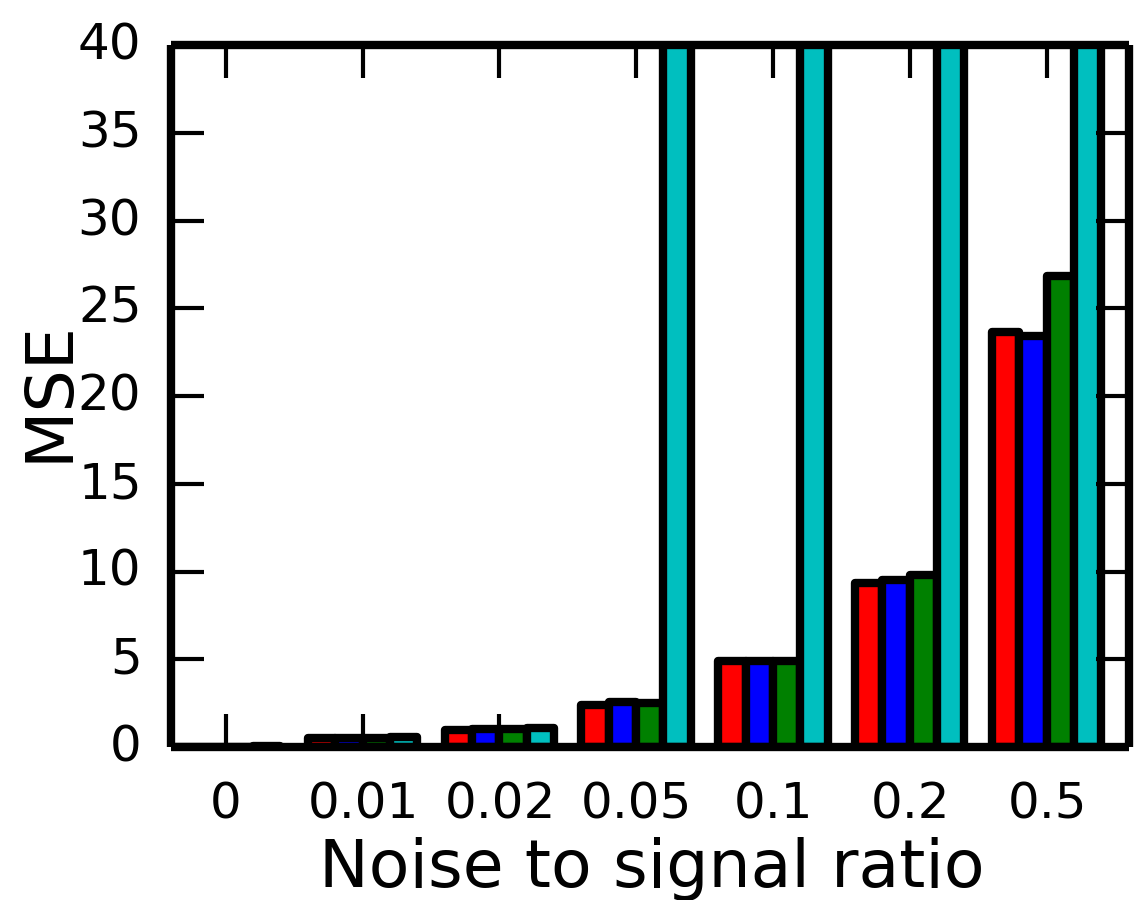}
		\captionsetup{width=0.9\columnwidth}
		\caption{NMF} 
		\label{mse_nmf_noise_test}
	\end{subfigure}
	\begin{subfigure}{0.245 \columnwidth}
		\includegraphics[width=\columnwidth]{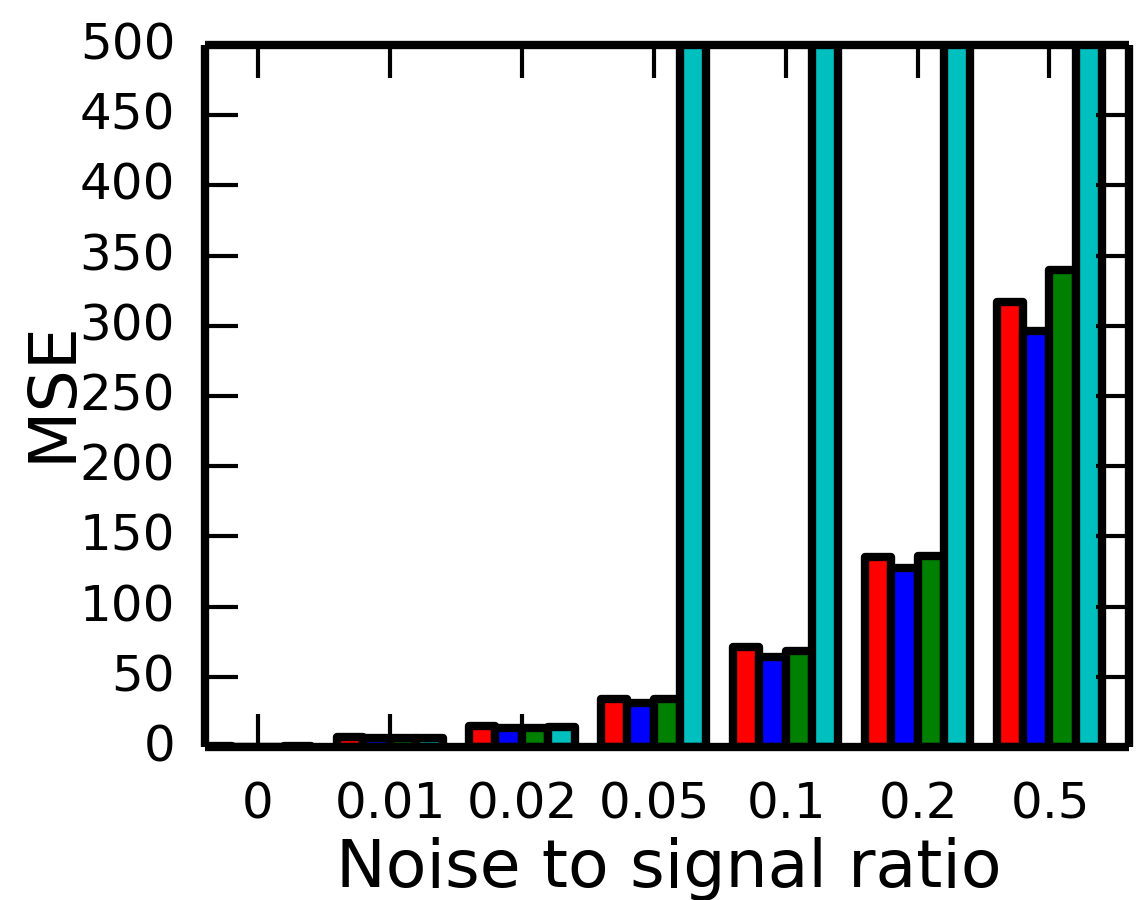}
		\captionsetup{width=0.9\columnwidth}
		\caption{NMTF} 
		\label{mse_nmtf_noise_test}
	\end{subfigure}
	\caption{Missing values prediction performances (Figures \ref{mse_nmf_missing_values_predictions} and \ref{mse_nmtf_missing_values_predictions}) and noise test performances (Figures \ref{mse_nmf_noise_test} and \ref{mse_nmtf_noise_test}), measured by average predictive performance on test set (mean square error) for different fractions of unknown values and noise-to-signal ratios.}
\end{figure*}

\subsection{Other experiments}
We conducted several other experiments, such as model selection on toy datasets, and cross-validation on three drug sensitivity datasets. Details and results for all experiments are given in the supplementary materials (Section 3), but here we highlight the results for missing value predictions and noise tests. In Figures \ref{mse_nmf_missing_values_predictions} and \ref{mse_nmtf_missing_values_predictions}, we measure the ability of the models to predict missing values in a toy dataset as the fraction of missing values increases. Similarly, Figures \ref{mse_nmf_noise_test} and \ref{mse_nmtf_noise_test} show the predictive performances as the amount of noise in the data increases.
The figures show that the Bayesian models are more robust to noise, and perform better on sparse datasets than their non-probabilistic counterparts.


\section{Conclusion}

We have introduced a fast variational Bayesian algorithm for performing non-negative matrix factorisation and tri-factorisation. We have shown that this method gives us deterministic convergence that is faster than MCMC methods, without requiring additional samples to estimate the posterior distribution. We demonstrate that our variational approach is particularly useful for the tri-factorisation case, where convergence is even harder, and we obtain a four-fold time speedup. These speedups can open up the applicability of the models to larger datasets.


\section*{Acknowledgement} 
This work was supported by the UK Engineering and Physical Sciences Research Council (EPSRC), grant reference EP/M506485/1. JF acknowledge funding from the Danish Council for Independent Research 0602-02909B.

\bibliography{bibliography}

\begin{thebibliography}{13}
\providecommand{\natexlab}[1]{#1}
\providecommand{\url}[1]{\texttt{#1}}
\expandafter\ifx\csname urlstyle\endcsname\relax
  \providecommand{\doi}[1]{doi: #1}\else
  \providecommand{\doi}{doi: \begingroup \urlstyle{rm}\Url}\fi

\bibitem[Ammad-ud din et~al.(2014)Ammad-ud din, Georgii, G\"{o}nen, Laitinen,
  Kallioniemi, Wennerberg, Poso, and Kaski]{Ammad-ud-din2014}
M.~Ammad-ud din, E.~Georgii, M.~G\"{o}nen, T.~Laitinen, O.~Kallioniemi,
  K.~Wennerberg, A.~Poso, and S.~Kaski.
\newblock {Integrative and personalized QSAR analysis in cancer by kernelized
  Bayesian matrix factorization.}
\newblock \emph{Journal of chemical information and modeling}, 54\penalty0
  (8):\penalty0 2347--59, Aug. 2014.

\bibitem[Barretina et~al.(2012)Barretina, Caponigro, Stransky, Venkatesan,
  Margolin, Kim, Wilson, et~al.]{Barretina2012}
J.~Barretina, G.~Caponigro, N.~Stransky, K.~Venkatesan, A.~A. Margolin, S.~Kim,
  C.~J. Wilson, et~al.
\newblock {The Cancer Cell Line Encyclopedia enables predictive modelling of
  anticancer drug sensitivity}.
\newblock \emph{Nature}, 483\penalty0 (7391):\penalty0 603--7, Mar. 2012.

\bibitem[Beal and Ghahramani(2003)]{J.M.Bernardo}
M.~Beal and Z.~Ghahramani.
\newblock {The Variational Bayesian EM Algorithm for Incomplete Data: with
  Application to Scoring Graphical Model Structures}.
\newblock \emph{Bayesian Statistics 7, Oxford University Press}, 2003.

\bibitem[Chen et~al.(2009)Chen, Wang, and Zhang]{Chen2009}
G.~Chen, F.~Wang, and C.~Zhang.
\newblock {Collaborative filtering using orthogonal nonnegative matrix
  tri-factorization}.
\newblock \emph{Information Processing and Management}, 45\penalty0
  (3):\penalty0 368--379, May 2009.

\bibitem[Ding et~al.(2006)Ding, Li, Peng, and Park]{Ding2006}
C.~Ding, T.~Li, W.~Peng, and H.~Park.
\newblock {Orthogonal nonnegative matrix t-factorizations for clustering}.
\newblock In \emph{Proceedings of the 12th ACM SIGKDD}, pages 126--135, New
  York, New York, USA, Aug. 2006. ACM Press.

\bibitem[Hwang et~al.(2012)Hwang, Atluri, Xie, Dey, Hong, Kumar, and
  Kuang]{Hwang2012}
T.~Hwang, G.~Atluri, M.~Xie, S.~Dey, C.~Hong, V.~Kumar, and R.~Kuang.
\newblock {Co-clustering phenome-genome for phenotype classification and
  disease gene discovery}.
\newblock \emph{Nucleic Acids Research}, 40\penalty0 (19):\penalty0 e146, Oct.
  2012.

\bibitem[Lee and Seung(1999)]{Lee1999}
D.~D. Lee and H.~S. Seung.
\newblock {Learning the parts of objects by non-negative matrix factorization.}
\newblock \emph{Nature}, 401\penalty0 (6755):\penalty0 788--791, Oct. 1999.

\bibitem[Lee and Seung(2000)]{Lee2000}
D.~D. Lee and H.~S. Seung.
\newblock {Algorithms for Non-negative Matrix Factorization}.
\newblock \emph{NIPS, MIT Press}, pages 556--562, 2000.

\bibitem[Schmidt et~al.(2009)Schmidt, Winther, and Hansen]{Schmidt2009}
M.~N. Schmidt, O.~Winther, and L.~K. Hansen.
\newblock {Bayesian non-negative matrix factorization}.
\newblock In \emph{Independent Component Analysis and Signal Separation,
  International Conference on (ICA), Springer Lecture Notes in Computer
  Science, Vol. 5441}, pages 540--547, 2009.

\bibitem[Seashore-Ludlow et~al.(2015)Seashore-Ludlow, Rees, Cheah, Cokol,
  Price, Coletti, Jones, et~al.]{Seashore-Ludlow2015}
B.~Seashore-Ludlow, M.~G. Rees, J.~H. Cheah, M.~Cokol, E.~V. Price, M.~E.
  Coletti, V.~Jones, et~al.
\newblock {Harnessing Connectivity in a Large-Scale Small-Molecule Sensitivity
  Dataset.}
\newblock \emph{Cancer discovery}, 5\penalty0 (11):\penalty0 1210--23, Nov.
  2015.

\bibitem[Wang et~al.(2013)Wang, Wang, and Gao]{Wang2013}
J.~J.-Y. Wang, X.~Wang, and X.~Gao.
\newblock {Non-negative matrix factorization by maximizing correntropy for
  cancer clustering.}
\newblock \emph{BMC bioinformatics}, 14\penalty0 (1):\penalty0 107, Jan. 2013.

\bibitem[Yang et~al.(2013)Yang, Soares, Greninger, Edelman, Lightfoot, Forbes,
  Bindal, Beare, Smith, Thompson, Ramaswamy, Futreal, Haber, Stratton, Benes,
  McDermott, and Garnett]{Yang2013}
W.~Yang, J.~Soares, P.~Greninger, E.~J. Edelman, H.~Lightfoot, S.~Forbes,
  N.~Bindal, D.~Beare, J.~A. Smith, I.~R. Thompson, S.~Ramaswamy, P.~A.
  Futreal, D.~A. Haber, M.~R. Stratton, C.~Benes, U.~McDermott, and M.~J.
  Garnett.
\newblock {Genomics of Drug Sensitivity in Cancer (GDSC): a resource for
  therapeutic biomarker discovery in cancer cells.}
\newblock \emph{Nucleic acids research}, 41\penalty0 (Database issue):\penalty0
  D955--61, Jan. 2013.

\bibitem[Yoo and Choi(2009)]{Yoo2009}
J.~Yoo and S.~Choi.
\newblock {Probabilistic matrix tri-factorization}.
\newblock In \emph{IEEE International Conference on Acoustics, Speech, and
  Signal Processing}, number~3, pages 1553--1556. IEEE, Apr. 2009.

\end{thebibliography}


\begin{thebibliography}{9}
\providecommand{\natexlab}[1]{#1}
\providecommand{\url}[1]{\texttt{#1}}
\expandafter\ifx\csname urlstyle\endcsname\relax
  \providecommand{\doi}[1]{doi: #1}\else
  \providecommand{\doi}{doi: \begingroup \urlstyle{rm}\Url}\fi

\bibitem[Akaike(1974)]{Akaike1974}
H.~Akaike.
\newblock {A new look at the statistical model identification}.
\newblock \emph{IEEE Transactions on Automatic Control}, 19\penalty0
  (6):\penalty0 716--723, Dec. 1974.

\bibitem[Ammad-ud din et~al.(2014)Ammad-ud din, Georgii, G\"{o}nen, Laitinen,
  Kallioniemi, Wennerberg, Poso, and Kaski]{Ammad-ud-din2014}
M.~Ammad-ud din, E.~Georgii, M.~G\"{o}nen, T.~Laitinen, O.~Kallioniemi,
  K.~Wennerberg, A.~Poso, and S.~Kaski.
\newblock {Integrative and personalized QSAR analysis in cancer by kernelized
  Bayesian matrix factorization.}
\newblock \emph{Journal of chemical information and modeling}, 54\penalty0
  (8):\penalty0 2347--59, Aug. 2014.

\bibitem[Barretina et~al.(2012)Barretina, Caponigro, Stransky, Venkatesan,
  Margolin, Kim, Wilson, et~al.]{Barretina2012}
J.~Barretina, G.~Caponigro, N.~Stransky, K.~Venkatesan, A.~A. Margolin, S.~Kim,
  C.~J. Wilson, et~al.
\newblock {The Cancer Cell Line Encyclopedia enables predictive modelling of
  anticancer drug sensitivity}.
\newblock \emph{Nature}, 483\penalty0 (7391):\penalty0 603--7, Mar. 2012.

\bibitem[Ding et~al.(2006)Ding, Li, Peng, and Park]{Ding2006}
C.~Ding, T.~Li, W.~Peng, and H.~Park.
\newblock {Orthogonal nonnegative matrix t-factorizations for clustering}.
\newblock In \emph{Proceedings of the 12th ACM SIGKDD}, pages 126--135, New
  York, New York, USA, Aug. 2006. ACM Press.

\bibitem[Lee and Seung(2000)]{Lee2000}
D.~D. Lee and H.~S. Seung.
\newblock {Algorithms for Non-negative Matrix Factorization}.
\newblock \emph{NIPS, MIT Press}, pages 556--562, 2000.

\bibitem[Schmidt et~al.(2009)Schmidt, Winther, and Hansen]{Schmidt2009}
M.~N. Schmidt, O.~Winther, and L.~K. Hansen.
\newblock {Bayesian non-negative matrix factorization}.
\newblock In \emph{Independent Component Analysis and Signal Separation,
  International Conference on (ICA), Springer Lecture Notes in Computer
  Science, Vol. 5441}, pages 540--547, 2009.

\bibitem[Schwarz(1978)]{Schwarz1978}
G.~Schwarz.
\newblock {Estimating the Dimension of a Model}.
\newblock \emph{The Annals of Statistics}, 6\penalty0 (2):\penalty0 461--464,
  Mar. 1978.

\bibitem[Yang et~al.(2013)Yang, Soares, Greninger, Edelman, Lightfoot, Forbes,
  Bindal, Beare, Smith, Thompson, Ramaswamy, Futreal, Haber, Stratton, Benes,
  McDermott, and Garnett]{Yang2013}
W.~Yang, J.~Soares, P.~Greninger, E.~J. Edelman, H.~Lightfoot, S.~Forbes,
  N.~Bindal, D.~Beare, J.~A. Smith, I.~R. Thompson, S.~Ramaswamy, P.~A.
  Futreal, D.~A. Haber, M.~R. Stratton, C.~Benes, U.~McDermott, and M.~J.
  Garnett.
\newblock {Genomics of Drug Sensitivity in Cancer (GDSC): a resource for
  therapeutic biomarker discovery in cancer cells.}
\newblock \emph{Nucleic acids research}, 41\penalty0 (Database issue):\penalty0
  D955--61, Jan. 2013.

\bibitem[Yoo and Choi(2009)]{Yoo2009}
J.~Yoo and S.~Choi.
\newblock {Probabilistic matrix tri-factorization}.
\newblock In \emph{IEEE International Conference on Acoustics, Speech, and
  Signal Processing}, number~3, pages 1553--1556. IEEE, Apr. 2009.

\end{thebibliography}
\bibliographystyle{abbrvnat}

\end{document}


\title{Fast Bayesian Non-Negative Matrix Factorisation and Tri-Factorisation \\ \vspace{10pt} Supplementary Materials \\ \vspace{20pt} \large{Thomas Brouwer, Jes Frellsen, Pietro Lio' \\ University of Cambridge}}
	\author{}
	\date{}
	\maketitle{}
	
	\noindent \large{Submitted to NIPS 2016 Workshop -- Advances in Approximate Bayesian Inference.} \\
	
	\tableofcontents
	
	\newpage
		

\section{Model details}

	\subsection{Gibbs sampling for matrix factorisation} \label{GibbsNMF}
		In this section we offer an introduction to Gibbs sampling, and show how it can be applied to the Bayesian non-negative matrix factorisation model. \\
		
		\noindent Gibbs sampling works by sampling new values for each parameter $ \theta_i $ from its marginal distribution given the current values of the other parameters $ \btheta_{-i} $, and the observed data $ D $. If we sample new values in turn for each parameter $ \theta_i $ from $ p(\theta_i | \btheta_{-i}, D ) $, we will eventually converge to draws from the posterior, which can be used to approximate the posterior $ p(\btheta|D) $. We have to discard the first $n$ draws because it takes a while to converge (\textit{burn-in}), and since consecutive draws are correlated we only use every $i$th value (\textit{thinning}). \\ 
		
		\noindent For the Bayesian non-negative matrix factorisation model this means that we need to be able to draw from the following distributions: 
		\begin{align*}
			p(U_{ik}|\tau,\U_{-ik},\V,D)	\quad\quad\quad		p(V_{jk}|\tau,\U,\V_{-jk},D)		\quad\quad\quad 	p(\tau|\U,\V,D)
		\end{align*}
		where $\U_{-ik}$ denotes all elements in $\U$ except $U_{ik}$, and similarly for $\V_{-jk}$. Using Bayes theorem we can obtain the posterior distributions. For example, for $p(U_{ik}|\tau,\U_{-ik},\V,D)$:
		\begin{alignat*}{1}
			p(U_{ik}|\tau,\U_{-ik},\V,D) &\propto p(D|\tau,\U,\V) \times p(U_{ik}|\lambdaUik) \\
			&\propto \prod_{j \in \Omega_i} \mathcal{N} (R_{ij} | \U_i \cdot \V_j, \tau^{-1} ) \times \mathcal{E} ( U_{ik} | \lambdaUik) \\
			&\propto \exp \left\{ \frac{\tau}{2} \sumOmegai (R_{ij} - \U_i \cdot \V_j)^2 \right\} \times \exp \left\{ - \lambdaUik U_{ik} \right\} \times u(x) \\
			&\propto \exp \left\{ \frac{U_{ik}^2}{2} \left[ \displaystyle \tau \sumOmegai V_{jk}^2 \right] + U_{ik} \left[ - \lambdaUik + \tau \sumOmegai \diffexclk V_{jk} \right] \right\} \times u(x) \\
			&\propto \exp \left\{ \frac{\tauUik}{2} ( U_{ik} - \muUik )^2 \right\} \times u(x) \\
			&\propto \mathcal{TN} ( U_{ik} | \muUik, \tauUik )
		\end{alignat*}
		where 
		\begin{equation*}
			\mathcal{TN} ( x | \mu, \tau ) = \left\{
			\begin{array}{ll}
				\displaystyle \frac{ \sqrt{ \frac{\tau}{2\pi} } \exp \left\{ -\frac{\tau}{2} (x - \mu)^2 \right\} }{ 1 - \Phi ( - \mu \sqrt{\tau} )}  & \mbox{if } x \geq 0 \\
				0 & \mbox{if } x < 0
			\end{array}
			\right.
		\end{equation*}
		is a truncated normal: a normal distribution with zero density below $ x = 0 $ and renormalised to integrate to one. $ \Phi(\cdot) $ is the cumulative distribution function of $ \mathcal{N}(0,1) $. \\
		
		\noindent Applying the same technique to the other posteriors gives us:
		\begin{alignat*}{1}
			p(\tau|\U,\V,D) &= \mathcal{G} (\tau | \alpha^*, \beta^* ) \\
			p(V_{jk}|\tau,\U,\V_{-jk},D) &= \mathcal{TN} ( V_{jk} | \muVjk, \tauVjk )
		\end{alignat*}
		The parameters of these distributions are given in Table \ref{bnmf_updates}, where $ \Omega_i = \left\{ j \text{ $ \vert $ } (i,j) \in \Omega \right\} $ and  $ \Omega_j = \left\{ i \text{ $ \vert $ } (i,j) \in \Omega \right\} $. 
		
	\subsection{Variational Bayes for matrix factorisation}
		For the Variational Bayes algorithm for inference, updates for the approximate posterior distributions are given in Table \ref{bnmf_updates}, and were obtained using the techniques described in the paper.
		We use $ \widetilde{f(X)} $ as a shorthand for $ \mathbb{E}_q \left[ f(X) \right] $, where $X$ is a random variable and $f$ is a function over $X$. We make use of the identity $ \widetilde{X^2} = \widetilde{X}^2 + \mathrm{Var}_q \left[ X \right] $.
		The expectation and variance of the parameters with respect to $ q $ are given below, for random variables $ X \sim \mathcal{G}(a,b) $ and $ Y \sim \mathcal{TN}(\mu,\tau) $. 
		\begin{align*}
			\widetilde{X} = \frac{a}{b}		\quad\quad\quad 		\widetilde{Y} = \mu + \frac{1}{\sqrt{\tau}} \lambda \left( - \mu \sqrt{ \tau } \right)	\quad\quad\quad		\mathrm{Var} \left[ Y \right] = \frac{1}{\tau} \left[ 1 - \delta \left( - \mu \sqrt{ \tau } \right) \right]
		\end{align*}
		\noindent where $ \psi(x) = \frac{d}{dx} \log \Gamma(x) $ is the digamma function, $ \lambda(x) = \phi(x) / [ 1 - \Phi(x) ] $, and $ \delta(x) = \lambda(x) [ \lambda(x) - x ] $. $ \phi(x) = \frac{1}{\sqrt{2\pi}} \exp \lbrace - \frac{1}{2} x^2 \rbrace $ is the density function of $ \mathcal{N}(0,1) $. 

		\begin{table*}[tp]
			\caption{NMF variable update rules} \label{bnmf_updates}
			\begin{center}
				\begin{tabular}{c|l|l}
					{\bf}  &{\bf GIBBS SAMPLING}  &{\bf VARIATIONAL BAYES} \\
					\hline & \\
					$ \alpha^* $ 		& $ \displaystyle \alpha + \frac{|\Omega|}{2} $		& $ \alpha + \frac{|\Omega|}{2}	$	\\
					\rule{0pt}{3ex}  
					$ \beta^* $			& $ \displaystyle \beta + \frac{1}{2} \sumOmega (R_{ij} - \U_i \cdot \V_j)^2 $ 		& $ \beta + \frac{1}{2} \sumOmega \expdiff $ \\
					\rule{0pt}{3ex}  
					$\tauUik $			& $ \displaystyle \tau \sumOmegai V_{jk}^2$ 		& $ \exptau \sumOmegai \expVjksqr $		\\	
					\rule{0pt}{3ex}
					$ \muUik $			& $ \displaystyle \frac{1}{\tauUik} \left(  - \lambdaUik + \tau \sumOmegai \diffexclk V_{jk} \right) $ 		& $ \frac{1}{\tauUik} \left( - \lambdaUik + \exptau \sumOmegai \diffexpexclk \expVjk \right) $ \\
					\rule{0pt}{3ex}
					$ \tauVjk $			& $ \displaystyle \tau \sumOmegaj U_{ik}^2 $ 		& $ \tau \sumOmegaj \expUiksqr $ 	\\
					\rule{0pt}{3ex}
					$ \muVjk $			& $ \displaystyle \frac{1}{\tauVjk} \left(  - \lambdaVjk + \tau \sumOmegaj \diffexclk U_{ik} \right) $ 			& $ \frac{1}{\tauVjk} \left(  - \lambdaVjk + \exptau \sumOmegaj \diffexpexclk \expUik \right) $ \\
					& \\
					\hline 
					\multicolumn{3}{c}{} \\
					\multicolumn{3}{c}{ $ \displaystyle \expdiff = \diffexp^2 + \sum_{k=1}^{K} \left( \expUiksqr \expVjksqr - \expUik^2 \expVjk^2 \right) $ } \\ 
					\multicolumn{3}{c}{} \\
					\hline 
				\end{tabular}
			\end{center}
		\end{table*}
		
	\subsection{BNMTF Gibbs sampling parameter values}
		For the BNMTF Gibbs sampling algorithm, we sample from the following posteriors:
		\begin{alignat*}{1}
			p(\tau|\F,\S,\G,D) &= \mathcal{G} (\tau | \alpha^*, \beta^* ) \\
			p(F_{ik}|\tau,\F_{-ik},\S,\G,D) &= \mathcal{TN} ( F_{ik} | \muFik, \tauFik ) \\
			p(S_{kl}|\tau,\F,\S_{-kl},\G,D) &= \mathcal{TN} ( S_{kl} | \muSkl, \tauSkl ) \\
			p(G_{jl}|\tau,\F,\S,\G_{-jl},D) &= \mathcal{TN} ( G_{jl} | \muGjl, \tauGjl ) 
		\end{alignat*}
		The updates for the parameters are given in Table \ref{bnmtf_gibbs_updates} below.
		\begin{table*}[h]
			\caption{NMTF Gibbs Update Rules} \label{bnmtf_gibbs_updates}
			\begin{center}
				\begin{tabular}{c|l}
					{\bf}  &{\bf GIBBS SAMPLING} \\
					\hline & \\
					$ \alpha^* $ 		& $ \displaystyle \alpha + \frac{|\Omega|}{2} $	\\
					$ \beta^* $			& $ \displaystyle \beta + \frac{1}{2} \sumOmega\diffTRI^2 $	 \\
					$ \tauFik $			& $ \displaystyle \tau \sumOmegai \left( \S_k \cdot \G_j \right)^2 $		\\
					$ \muFik $			& $ \displaystyle \frac{1}{\tauFik} \left(  - \lambdaFik + \tau \sumOmegai \diffTRIexclK \left( \S_k \cdot \G_j \right) \right) $ \\	
					$ \tauSkl $			& $ \displaystyle \tau \sumOmega F_{ik}^2 G_{jl}^2 $		\\
					$ \muSkl $			& $ \displaystyle \frac{1}{\tauSkl} \left( - \lambdaSkl + \tau \sumOmega \diffTRIexclKL F_{ik} G_{jl} \right) $ \\
					$ \tauGjl $			& $ \displaystyle \tau \sumOmegaj \left( \F_i \cdot \S_{\cdot,l} \right)^2 $ \\
					$ \muGjl $			& $ \displaystyle \frac{1}{\tauGjl} \left( - \lambdaGjl + \tau \sumOmegaj \diffTRIexclL \left( \F_i \cdot \S_{\cdot,l} \right) \right) $ \\ \\
					\hline
				\end{tabular}
			\end{center}
		\end{table*}
	
	\subsection{BNMTF Variational Bayes parameter updates}
		As discussed in the paper, the term $ \expdiffTRI $ adds extra complexity to the matrix tri-factorisation case. 
		\begin{alignat}{1}
			\nonumber &\expdiffTRI = \\
			\nonumber &	\quad\quad	\diffexpTRI^2 \\
			&	\quad\quad	+ \sumk \suml \mathrm{Var}_q \left[ F_{ik} S_{kl} G_{jl} \right] \\
			&	\quad\quad	+ \sumk \suml \sumexclk \mathrm{Cov} \left[ F_{ik} S_{kl} G_{jl}, F_{ik'} S_{k'l} G_{jl} \right] \\
			&	\quad\quad	+ \sumk \suml \sumexcll \mathrm{Cov} \left[ F_{ik} S_{kl} G_{jl}, F_{ik} S_{kl'} G_{jl'} \right]
		\end{alignat}
		The above variance and covariance terms are equal to the following, respectively:
		\setcounter{equation}{0}
		\begin{alignat}{1}
			& \expFiksqr \expSklsqr \expGjlsqr - \expFik^2 \expSkl^2 \expGjl^2 \\
			& \varFik \expSkl \expGjl \expSklp \expGjlp \\	
			& \expFik \expSkl \varGjl \expFikp \expSkpl	
		\end{alignat}
		The updates for the variational parameters of the Variational Bayes algorithm for the Bayesian non-negative matrix tri-factorisation are given in Table \ref{bnmtf_vb_updates} below. \\
		
		\begin{table*}[h]
			\caption{NMTF VB Update Rules} \label{bnmtf_vb_updates}
			\begin{center}
				\begin{tabular}{c|l}
					{\bf}  &{\bf VARIATIONAL BAYES} \\
					\hline \\
					$ \alpha^* $ 		& $ \displaystyle \alpha + \frac{|\Omega|}{2} $	\\
					$ \beta^* $			& $ \displaystyle \beta + \frac{1}{2} \sumOmega \expdiffTRI $	 \\
					$ \tauFik $			& $ \displaystyle \exptau \sumOmegai \left( \left( \sum_{l=1}^L \expSkl \expGjl \right)^2 + \sum_{l=1}^L \left( \expSklsqr \expGjlsqr - \expSkl^2 \expGjl^2 \right) \right) $		\\
					$ \muFik $			& $ \displaystyle \frac{1}{\tauFik} \left( - \lambdaFik + \exptau \sumOmegai \left( \diffexpTRIexclK \sum_{l=1}^L \expSkl \expGjl - \suml \expSkl \varGjl \sumexclk \expFikp \expSkpl \right) \right) $ \\	
					$ \tauSkl $			& $ \displaystyle \exptau \sumOmega \expFiksqr \expGjlsqr $		\\
					$ \muSkl $			& $ \displaystyle \frac{1}{\tauSkl} \left( - \lambdaSkl + \exptau \sumOmega \left( \diffexpTRIexclKL \expFik \expGjl \right. \right. $ \\
										& \hspace{103pt} $ \displaystyle \left. \left. - \expFik \varGjl \sumexclk \expFikp \expSkpl  - \varFik \expGjl \sumexcll \expSklp \expGjlp \right) \right) $ \\
					$ \tauGjl $			& $ \displaystyle \exptau \sumOmegaj \left( \left( \sum_{k=1}^K \expFik \expSkl \right)^2 + \sum_{k=1}^K \left( \expFiksqr \expSklsqr - \expFik^2 \expSkl^2 \right) \right) $ \\
					$ \muGjl $			& $ \displaystyle \frac{1}{\tauGjl} \left( - \lambdaGjl + \exptau \sumOmegaj \left( \diffexpTRIexclL \sum_{k=1}^K \expFik \expSkl - \sumk \varFik \expSkl \sumexcll \expSklp \expGjlp \right) \right) $ \\ \\
					\hline
				\end{tabular}
			\end{center}
		\end{table*}
		
		
\newpage
\section{Model discussion}

	\subsection{Complexity}
		The updates for the Gibbs samplers and VB algorithms can be implemented efficiently using matrix operations. The time complexity per iteration for Bayesian non-negative matrix factorisation is $ \mathcal{O}( I J K^2 ) $ for both Gibbs and VB, and $ \mathcal{O}( I J (K^2 L + K L^2) ) $ per iteration for tri-factorisation. However, the updates in each column of $ \U, \V, \F, \G $ are independent of each other and can therefore be updated in parallel. \\
		
		\noindent For the Gibbs sampler, this means we can draw these values in parallel, but for the VB algorithm we can jointly update the columns using a single matrix operation. Modern computer architectures can exploit this using vector processors, leading to a great speedup. \\
		
		\noindent Furthermore, after the VB algorithm converges we have our approximation to the posterior distributions immediately, whereas with Gibbs we need to obtain further draws after convergence and use a thinning rate to obtain an accurate estimate of the posterior. This deterministic behaviour of VB makes it easier to use. Although additional variables need to be stored to represent the posteriors, this does not result in a worse space complexity, as the Gibbs sampler needs to store draws over time.
	
	\subsection{Model selection} \label{model_selection}
		In practice we do not know the optimal model dimensionality of our data, and we need to estimate its value. In our case, we want to find the best value of $ K $ for matrix factorisation, and $ K,L $ for tri-factorisation. \\
		
		\noindent The log probability of the data, $ \log p(D|\btheta) $, is a good measure for the quality of the fit to the data. As $ K $ increases we expect the log likelihood to improve as we give the model more freedom for fitting to the data, but this can lead to overfitting. Therefore, we need to penalise the model's performance by its complexity. We use the Akaike information criterion (AIC) \cite{Akaike1974} defined as 
		\begin{equation*} 
			AIC = 2k - 2 \log p(D|\btheta)
		\end{equation*} 
		where $ k $ is the number of free parameters in our model. For matrix factorisation this is $ IK + JK $ and for tri-factorisation $ IK + KL + JL $. \\
		
		\noindent Another popular measure is the Bayesian information criterion (BIC) \cite{Schwarz1978}.
		%
		%
		BIC tends to penalise complicated models more heavily than AIC. We found that AIC peaked closer to the true model dimensionality on synthetic data than BIC, especially for matrix tri-factorisation, and we therefore use the former. \\
		
\begin{figure}[h]
			\centering
			\begin{tikzpicture}
				\tikzstyle{main}=[circle, minimum size = 2mm, thick, draw =black, node distance = 8mm]
				\tikzstyle{connect}=[-latex, thick]
				\tikzstyle{box}=[rectangle, draw=black!100]
			
				\node[main, fill=black, label=below:{$(K_i,L_i)$}] (C1) {};
				\node[main, fill=black!40, label=below:{$(K_{i+1},L_i)$}, right=1.5cm of C1] (C2) {};
				\node[main, right=1.5cm of C2] (C3) {};
				\node[main, fill=black!40, label=above:{$(K_i,L_{i+1})$}, above=1.5cm of C1] (C4) {};
				\node[main, fill=black!40, label=above:{$(K_{i+1},L_{i+1})$}, right=1.5cm of C4] (C5) {};
				\node[main, right=1.5cm of C5] (C6) {};
				\node[main, above=1.5cm of C4] (C7) {};
				\node[main, right=1.5cm of C7] (C8) {};
				\node[main, right=1.5cm of C8] (C9) {};
				
			  	\path (C1) edge [connect, line width=0.4mm] (C2);
			  	\path (C1) edge [connect, line width=0.4mm] (C4);
			  	\path (C1) edge [connect, line width=0.4mm] (C5);
			\end{tikzpicture}
			\captionsetup{width=0.95\columnwidth}
			\caption{Greedy search procedure for model selection}
			\label{greedy_search}
		\end{figure}
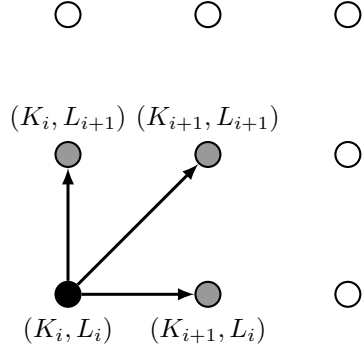

		\noindent For matrix factorisation we then try different values for $ K $ in a given range and pick the $ K $ that gives the lowest AIC. 
		Similarly for matrix tri-factorisation, we can perform a \textbf{grid search} for a range of values for $ K $ and $ L $, trying each possible $ (K,L) $ pair, but this results in training $ K \times L $ different models. Instead, we can perform a \textbf{greedy search} on the grid, as illustrated in Figure \ref{greedy_search}.
		\begin{itemize}
			\item We are given a grid of values $ (K_i,L_i) $.
			\item We start with the lowest values, $ (K_1,L_1) $, train a model, and measure the model quality.
			\item For each of the three points above it -- $ (K_i,L_{i+1})$, $(K_{i+1},L_i)$, $(K_{i+1},L_{i+1}) $ -- we train a model and measure the model quality.
			\item The model that gives the best improvement is selected as our next value on the grid. If no improvement is made, the current point $ (K_i,L_i) $ gives the best values for $ K $ and $ L $.
		\end{itemize}
		Since we are looking for the best fitting model, we can train multiple models with random initialisations for each $ K,L $ and use the one with the highest log likelihood (we denote \textit{restarts}).
	
	\subsection{Initialisation}
		Initialising the parameters of the models can vastly influence the quality of convergence. This can be done for example by using the hyperparameters $ \lambdaUik, \lambdaVjk, \lambdaFik, \lambdaSkl, \lambdaGjl, \alpha, \beta $ to set the initial values to the mean of the priors of the model. Alternatively, we can use random draws of the priors as the initial values. We found that random draws tend to give faster and better convergence than the expectation. \\
		
		\noindent For matrix tri-factorisation we can also initialise $ \F $ by running the K-means clustering algorithm on the rows as datapoints, and similarly $ \G $ for the columns, as suggested by \citet{Ding2006}. For the VB algorithm we then set the $ \mu $ parameters to the cluster indicators, and for Gibbs we set the matrices to the cluster indicators, plus $ 0.2 $ for smoothing. We found that this improved the convergence as well, with $ \S $ initialised using random draws.
		
	\subsection{Implementation}
		All algorithms mentioned were implemented using the Python language. The \textit{numpy} package was used for fast matrix operations, and for random draws of the truncated normal distribution we used the Python package \textit{rtnorm} by C. Lassner (\url{http://miv.u-strasbg.fr/mazet/rtnorm/}), giving more efficient draws than the standard libraries and dealing with rounding errors. \\
		
		\noindent The mean and variance of the truncated normal involve operations prone to numerical errors when $ \mu \ll 0 $. To deal with this we observe that when $ \mu \sqrt{\tau} \ll 0 $ the truncated normal distribution approximates an exponential one with rate $ \mu \sqrt{\tau} $, and therefore has mean $ 1 / ( \mu \sqrt{\tau} ) $ and variance $ 1 / (\mu \sqrt{\tau})^2 $. \\
		
		\noindent All experiments were run on a Medion Erazer laptop with an Intel i7-3610QM CPU (4 cores of 2.30GHz each), GeForce GTX 670M graphics card, and 16GB memory.
	
	\subsection{Code}
		Implementations of all discussed methods are available online, via \url{https://github.com/ThomasBrouwer/BNMTF/}.

		
\newpage
\section{Additional experiments}

	\subsection{Model selection}
		To demonstrate our proposed model selection framework (see section \ref{model_selection}) we use the toy dataset described earlier, using our VB algorithms. We let each model train for 1000 iterations with 5 restarts. \\
		
\begin{figure}[b!]
			\centering
			\captionsetup{width=0.9\columnwidth}
			\begin{subfigure}{0.25 \columnwidth}
				\includegraphics[width=\columnwidth]{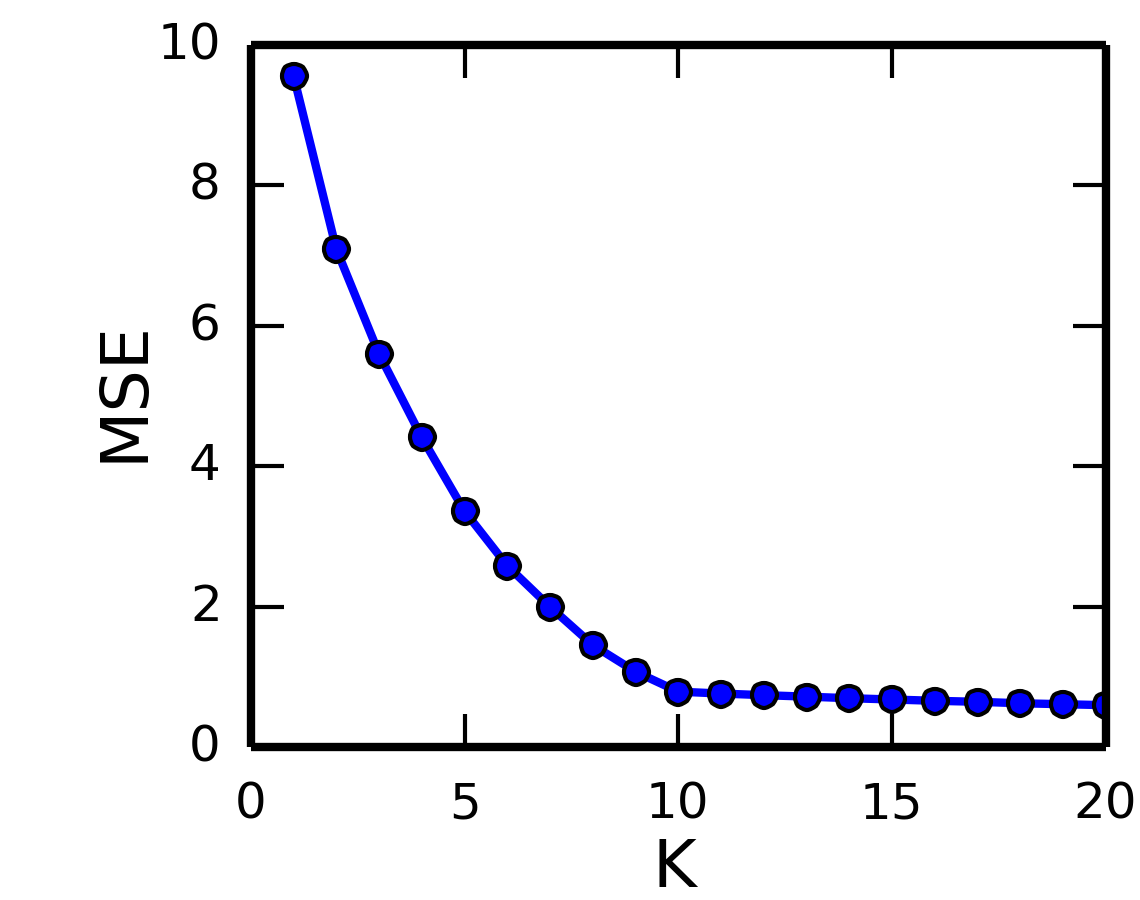}
				\captionsetup{width=0.9\columnwidth}
				\caption{MSE, line search} 
				\label{mse_nmf_model_selection}
			\end{subfigure}%
			\begin{subfigure}{0.25 \columnwidth}
				\includegraphics[width=\columnwidth]{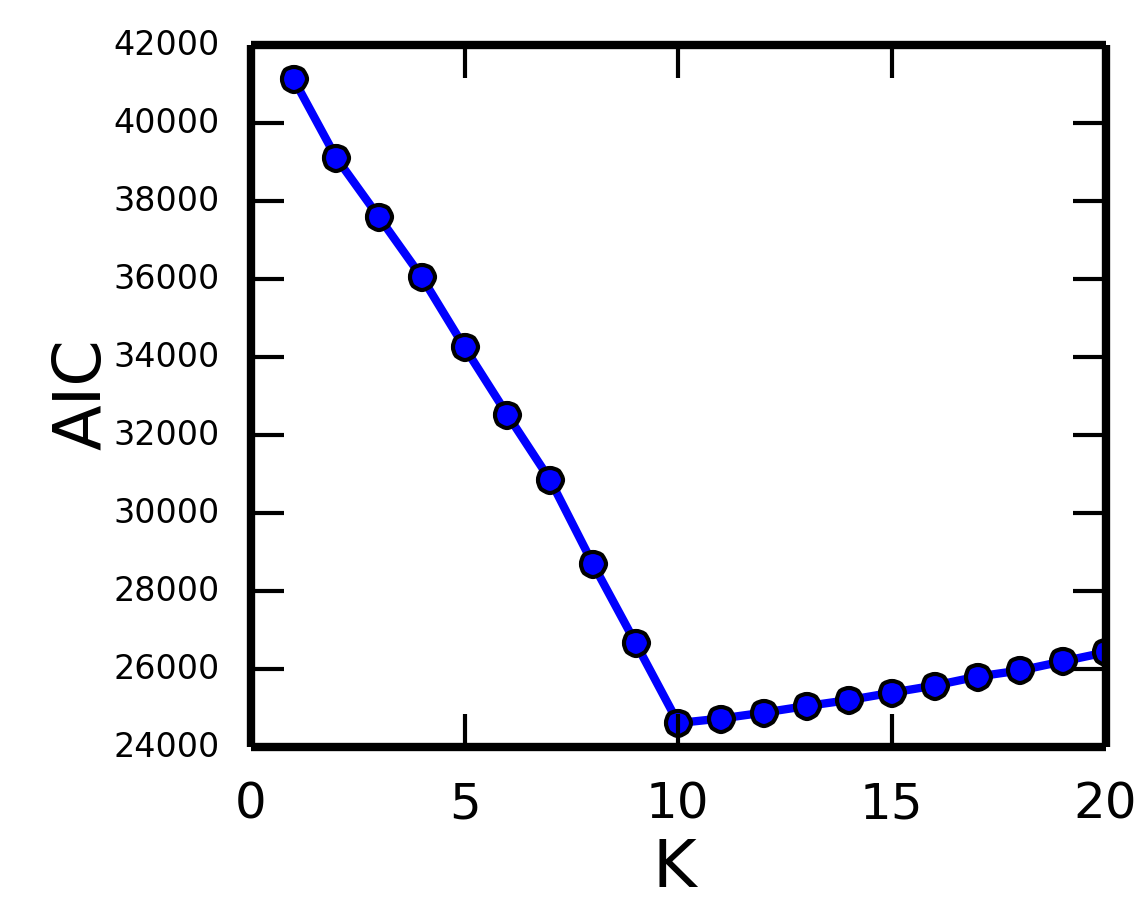}
				\captionsetup{width=0.9\columnwidth}
				\caption{AIC, line search} 
				\label{aic_nmf_model_selection}
			\end{subfigure}
			\vspace{10pt}
			\begin{subfigure}{0.25 \columnwidth}
				\includegraphics[width=\columnwidth]{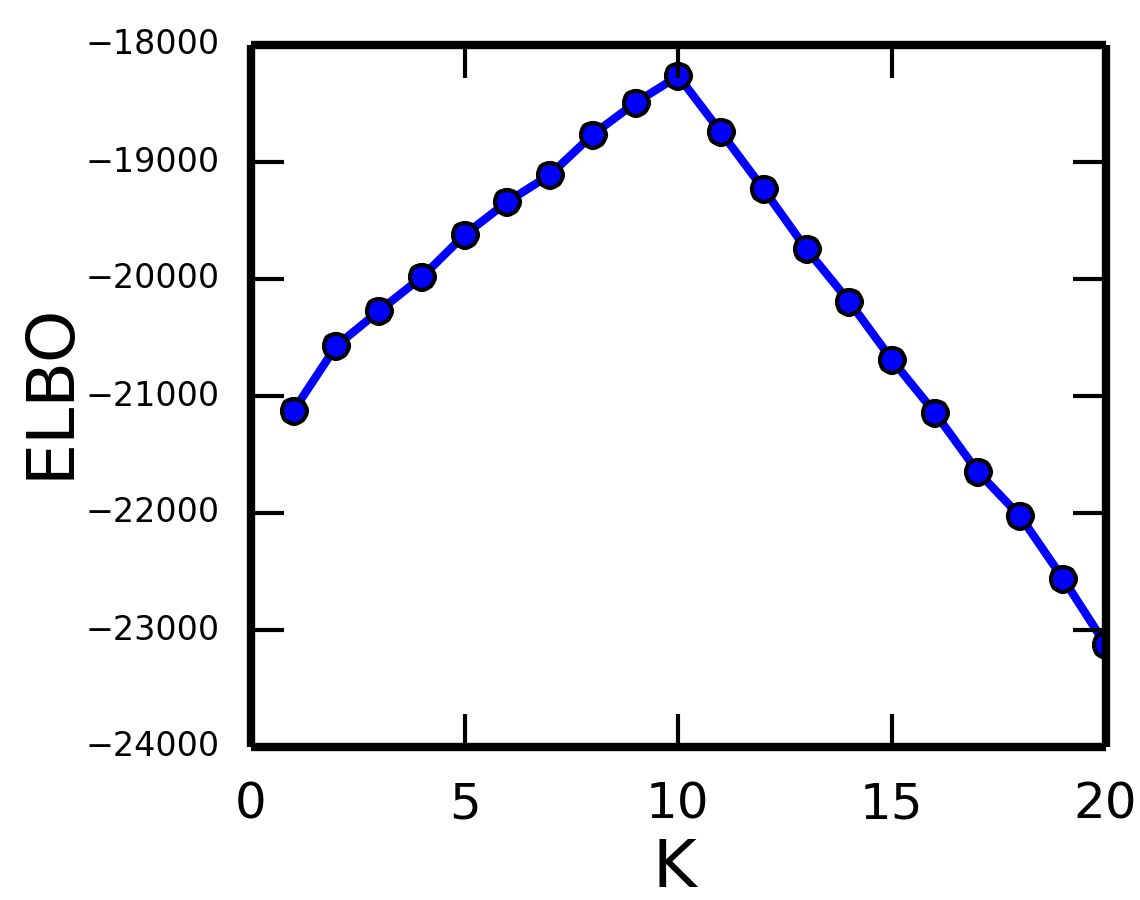}
				\captionsetup{width=0.9\columnwidth}
				\caption{ELBO, line search} 
				\label{elbo_nmf_model_selection}
			\end{subfigure}
			\begin{subfigure}{0.25 \columnwidth}
				\includegraphics[width=\columnwidth]{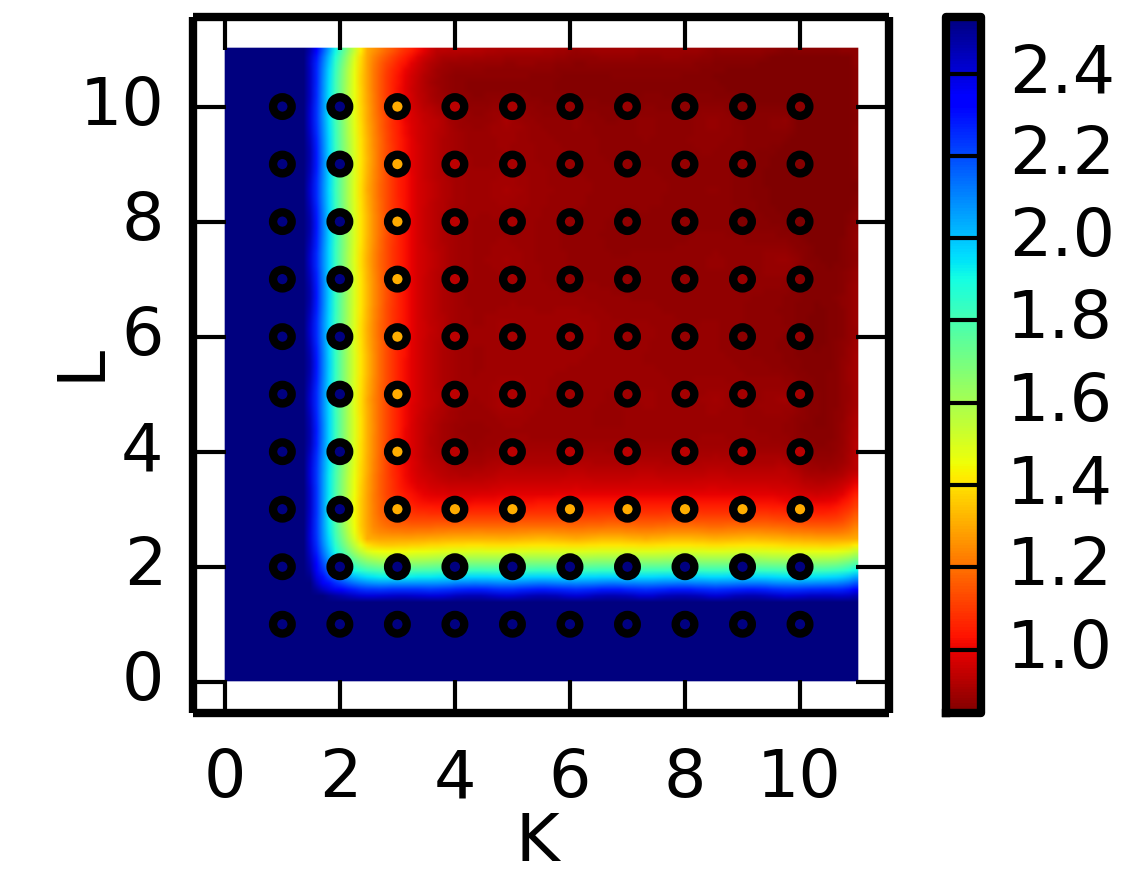}
				\captionsetup{width=0.9\columnwidth}
				\caption{MSE, grid search} 
				\label{mse_nmtf_model_selection}
			\end{subfigure}
			\begin{subfigure}{0.25 \columnwidth}
				\includegraphics[width=\columnwidth]{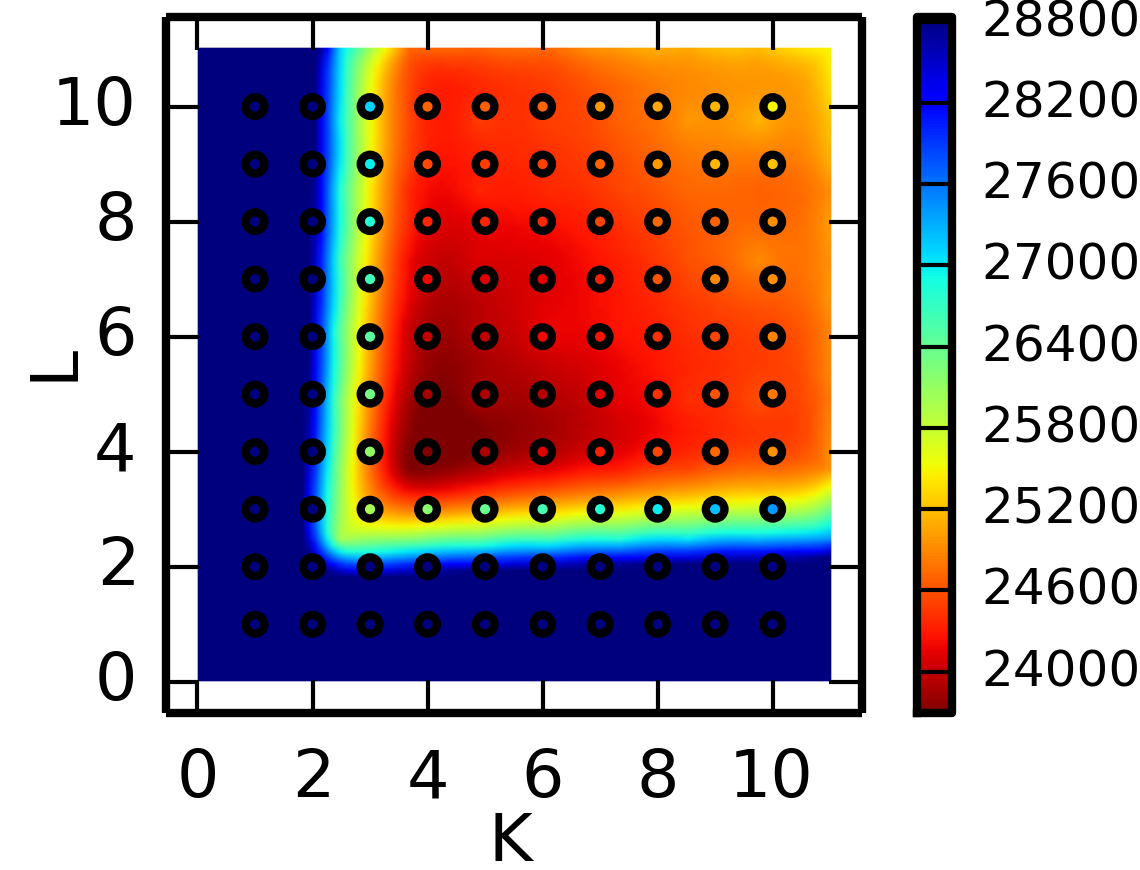}
				\captionsetup{width=0.9\columnwidth}
				\caption{AIC, grid search} 
				\label{aic_nmtf_model_selection}
			\end{subfigure}
			\begin{subfigure}{0.25 \columnwidth}
				\includegraphics[width=\columnwidth]{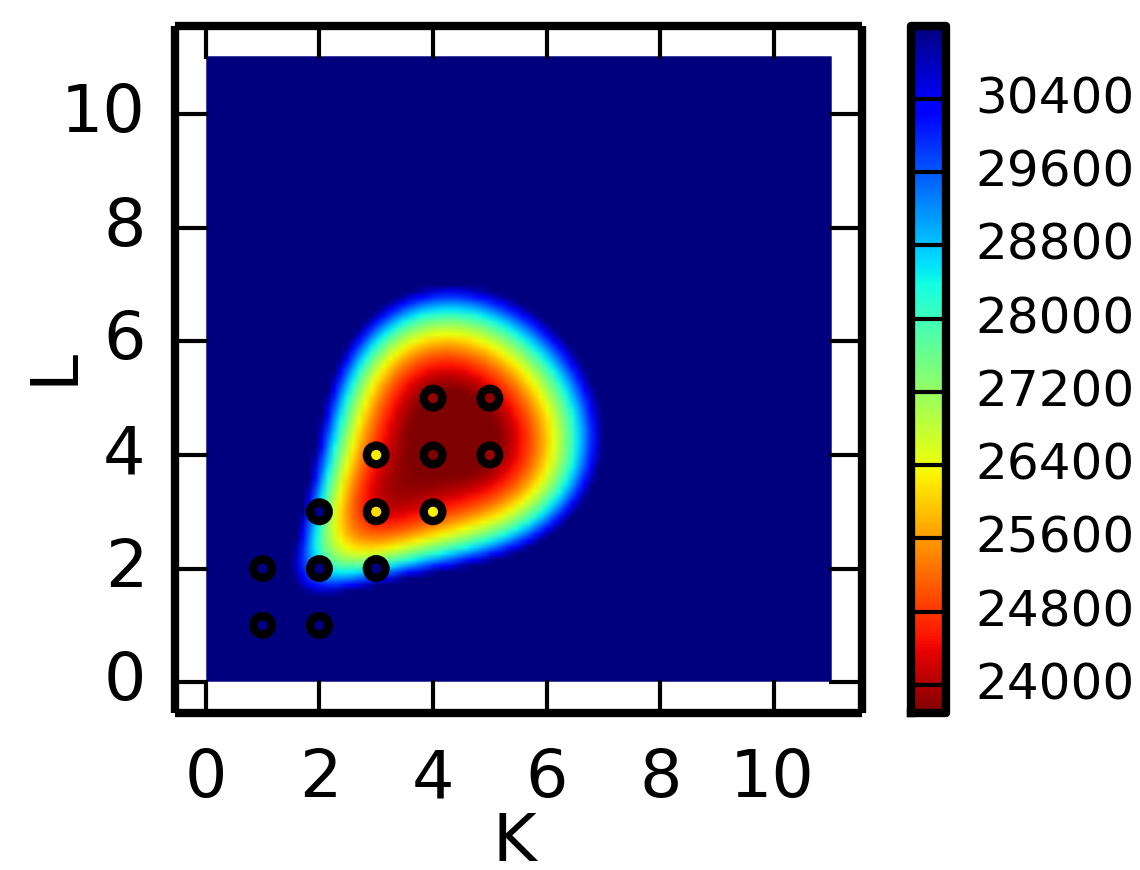}
				\captionsetup{width=0.9\columnwidth}
				\caption{AIC, greedy search} 
				\label{aic_nmtf_greedy_model_selection}
			\end{subfigure}
			\caption{Model selection for VB-NMF (top row) and VB-NMTF (bottom row). We measure the model quality for different values of $K$ (and $L$) on the toy datasets. The true $K$ for NMF is 10, and the true $K, L$ for NMTF is 5, 5. Figures \ref{mse_nmf_model_selection}--\ref{elbo_nmf_model_selection} show that the MSE cannot find the right model dimensionality for NMF, but the AIC and ELBO can. The same applies to NMTF, as shown in Figures \ref{mse_nmtf_model_selection}--\ref{aic_nmtf_greedy_model_selection}, where we additionally see that the proposed greedy search model selection method finds the same solution as the full grid one, but trying only 13 of the 100 possible values.}
			\label{model_selection_toy}
		\end{figure}

		\noindent As can be seen in Figure \ref{mse_nmf_model_selection}, the mean square error on the training data for matrix factorisation converges after $ K = 10 $, whereas Figure \ref{aic_nmf_model_selection} shows that using the Akaike information criterion gives a clear peak at the true $K$. The ELBO also provides a good heuristic for model selection, as seen in figure \ref{elbo_nmf_model_selection}. \\
		
		\noindent Figure \ref{aic_nmtf_model_selection} shows the full grid search for matrix tri-factorisaiton, and gives a peak at $ K = 4, L = 4 $. This is slightly lower than the true $ K, L $, but shows that a good enough fit can be achieved using fewer factors. The proposed greedy search in Figure \ref{aic_nmtf_greedy_model_selection} finds the same solution but only trying 13 of the 100 possible combinations, suggesting that this model selection procedure can offer a significant speedup with similar performance. \\
		
\begin{figure}[t!]
			\centering
			\captionsetup{width=0.9\columnwidth}
			\begin{subfigure}{0.25 \columnwidth}
				\includegraphics[width=\columnwidth]{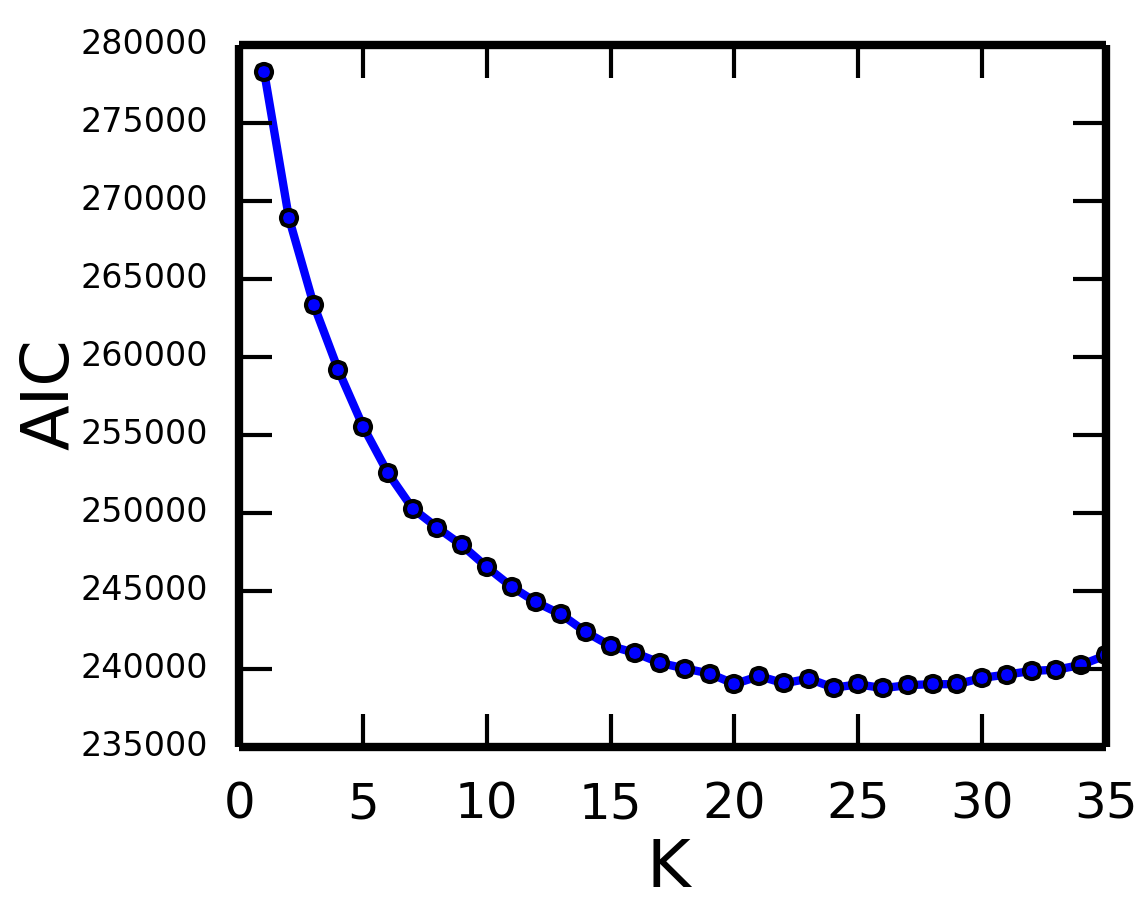}
				\captionsetup{width=0.9\columnwidth}
				\caption{AIC, line search} 
				\label{aic_Sanger_line_model_selection}
			\end{subfigure}
			\begin{subfigure}{0.25 \columnwidth}
				\includegraphics[width=\columnwidth]{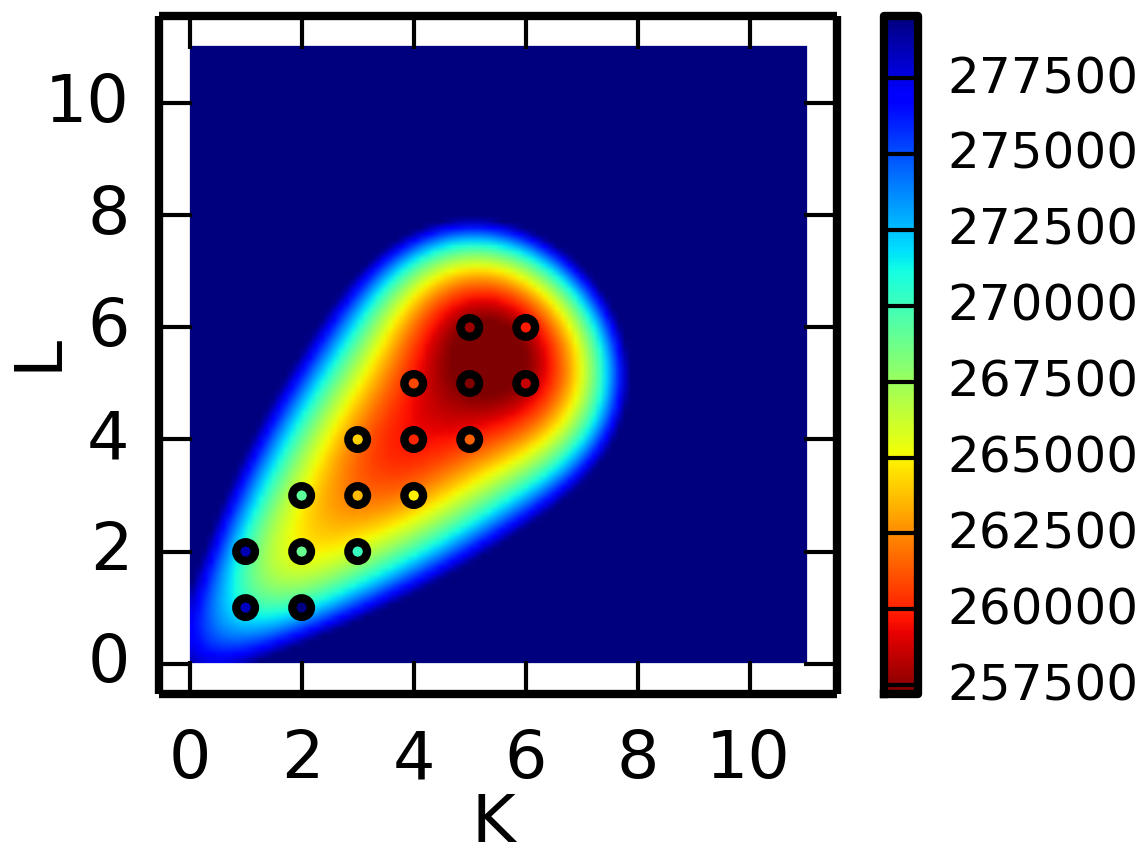}
				\captionsetup{width=0.9\columnwidth}
				\caption{AIC, greedy search} 
				\label{aic_Sanger_greedy_model_selection}
			\end{subfigure}
			\caption{Model selection for VB-NMF (top row) and VB-NMTF (bottom row) on the GDSC drug sensitivity dataset. We see that the found optimal model dimensionalities are $K=25$ for NMF, and $K=5, L=5$ for NMTF.}
			\label{model_selection_Sanger}
		\end{figure}

		\noindent We also ran the model selection frameworks on the drug sensitivity dataset, where the true number of latent factors are unknown. Figure \ref{model_selection_Sanger} shows that for matrix factorisation the best value for $ K $ is around $ 25 $, and for matrix tri-factorisation $ K = L = 5 $.
		
	\subsection{Missing values} 
		We furthermore tested the ability of our model to recover missing values as the fraction of unknown entries increases (more sparse datasets). We run each algorithm on the same dataset for 1000 iterations (burn-in 800, thinning rate 5) to give the algorithms enough time to converge, splitting the data randomly ten times each into test and training data, and computing the average mean square error of the predictions on the test data. \\
		
		\noindent High errors are indicative of overfitting or not converging to a good solution. We can see in Figure \ref{mse_nmf_missing_values_predictions} that the fully Bayesian methods for matrix factorisation obtain good predictive power even at 70\% missing values, whereas ICM starts failing there. The non-probabilistic method starts overfitting from 20\% missing values, leading to very high prediction errors. \\
		
		\noindent For matrix tri-factorisation we notice that our VB method sometimes does not converge to the best solution for 50\% or more missing values. This is shown in Figure \ref{mse_nmtf_missing_values_predictions}. As a result, the average error is higher than the other methods in those cases.
	
	\subsection{Noise test} 
\begin{figure*}[h]
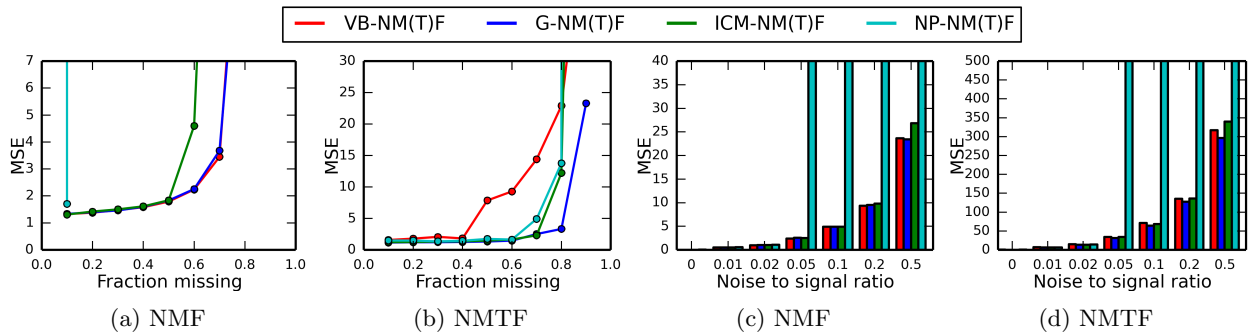

			\begin{subfigure}[t]{1\columnwidth}
				\hspace{100pt}
				\includegraphics[width=0.6\columnwidth]{legend.png}
			\end{subfigure}
			\\
			\captionsetup{width=0.9\columnwidth}
			\begin{subfigure}{0.245 \columnwidth}
				\includegraphics[width=\columnwidth]{mse_nmf_missing_values_predictions.png}
				\captionsetup{width=0.9\columnwidth}
				\caption{NMF} 
				\label{mse_nmf_missing_values_predictions}
			\end{subfigure} %
			\begin{subfigure}{0.245 \columnwidth}
				\includegraphics[width=\columnwidth]{mse_nmtf_missing_values_predictions.png}
				\captionsetup{width=0.9\columnwidth}
				\caption{NMTF} 
				\label{mse_nmtf_missing_values_predictions}
			\end{subfigure}
			\begin{subfigure}{0.245 \columnwidth}
				\includegraphics[width=\columnwidth]{mse_nmf_noise_test.png}
				\captionsetup{width=0.9\columnwidth}
				\caption{NMF} 
				\label{mse_nmf_noise_test}
			\end{subfigure}
			\begin{subfigure}{0.245 \columnwidth}
				\includegraphics[width=\columnwidth]{mse_nmtf_noise_test.png}
				\captionsetup{width=0.9\columnwidth}
				\caption{NMTF} 
				\label{mse_nmtf_noise_test}
			\end{subfigure}
			\caption{Missing values prediction performances (Figures \ref{mse_nmf_missing_values_predictions} and \ref{mse_nmtf_missing_values_predictions}) and noise test performances (Figures \ref{mse_nmf_noise_test} and \ref{mse_nmtf_noise_test}), measured by average predictive performance on test set (mean square error) for different fractions of unknown values and noise-to-signal ratios.}
		\end{figure*}

		We conducted a noise test to measure the robustness of the different methods. Our experiment works in a similar manner to the missing values test, but now adding different levels of Gaussian noise to the data, and with 10\% test data. The noise-to-signal ratio is given by the ratio of the variance of the Gaussian noise we add, to the variance of the generated data. We see in Figures \ref{mse_nmf_noise_test} and \ref{mse_nmtf_noise_test} that the non-probabilistic approach starts overfitting heavily at low levels of noise, whereas the fully Bayesian approaches achieve the best possible predictive powers even at high levels of noise.
		
	\subsection{Drug sensitivity predictions}
		Finally, we performed cross-validation experiments on three different drug sensitivity experiments. Firstly, the Genomics of Drug Sensitivity in Cancer (GDSC v5.0, \cite{Yang2013}) dataset contains 138 drugs and 622 cell lines, with 81\% of entries observed (as introduced in the paper). Secondly, the Cancer Cell Line Encyclopedia (CCLE, \cite{Barretina2012}) has 504 drugs and 22 cell lines. There are two versions: one detailing $IC_{50}$ drug sensitivity values (96\% observed) and another giving $EC_{50}$ values (63\% observed). \\
		
		\noindent We compare our methods against classic algorithms for matrix factorisation and tri-factorisation. Aside from the Gibbs sampler (G-NMF, G-NMTF) and VB algorithms (VB-NMF, VB-NMTF), we consider the non-probabilistic matrix factorisation (NP-NMF) and tri-factorisation (NP-NMTF) methods introduced by \citet{Lee2000} and \citet{Yoo2009}, respectively. \citet{Schmidt2009} also proposed an Iterated Conditional Modes (ICM-NMF) algorithm for computing an MAP solution, where instead of using draws from the posteriors as updates we set their values to the mode. We also extended this method for matrix tri-factorisation (ICM-NMTF). \\
		
		\noindent For the GDSC dataset we also compare with a recent paper by \citet{Ammad-ud-din2014} which uses a method called Kernelised Bayesian Matrix Factorisation (KBMF), leveraging similarity kernels of the drugs and cell lines. We reconstructed the drug kernels using targets, PaDeL fingerprints, and 1D and 2D descriptors. Similarly for the cell lines we used gene expression, copy-number variation, and cancer mutation data. For the other datasets we only compared the matrix factorisation models. \\
		
		\noindent The results of running 10-fold cross-validation can be found in Table \ref{cross_validation_table}. For KBMF and non-probabilistic NMF and NMTF we use nested cross-validation to find the best value for $ K $ (and $ L $). For the other methods we use cross-validation with the model selection detailed in the supplementary materials (Section \ref{model_selection}). \\
		
		\noindent We can see the Gibbs sampling NMF model performs the best in two of the three datasets, outperforming even the KBMF model which uses side information. The ICM models tend to overfit to the data, and often led to very high predictive errors. The non-probabilistic models do well on the large GDSC dataset, but less so on the small CCLE datasets with only 24 rows. The Bayesian models do significantly better on these two. \\
		
		\noindent The matrix tri-factorisation models generally perform as well as its matrix factorisation counterpart. For matrix factorisation, the fast VB version does worse than the Gibbs sampling variant. However, for matrix tri-factorisation VB outperforms Gibbs on two of the three datasets. 
		
		\begin{table}[t]
			\caption{10-fold cross-validation drug sensitivity prediction results (mean squared error). Very high predictive errors are replaced by $\infty$, and the best performances are highlighted in bold.} \label{cross_validation_table}
			\begin{center}
				\begin{tabular}{llllllllll}
					\toprule
					& & \multicolumn{4}{l}{NMF} & \multicolumn{4}{l}{NMTF} \\
					\cmidrule(lr){3-6} \cmidrule(lr){7-10}
					&	KBMF	&	NP	&	ICM	&	G	&	VB	&	NP	&	ICM	&	G	&	VB \\ 
					\midrule
					GDSC $IC_{50}$ & 2.144 & 2.251 & \multicolumn{1}{c}{$\infty$} & \textbf{2.055} & 2.282 & 2.258 & 2.199 & 2.402 & 2.321 \\
					CCLE $IC_{50}$ & \multicolumn{1}{c}{-} & 4.683 & \multicolumn{1}{c}{$\infty$} & \textbf{3.719} & 3.984 & 4.664 & 4.153 & 3.745 & 4.107 \\
					CCLE $EC_{50}$ & \multicolumn{1}{c}{-} & 8.047 & \multicolumn{1}{c}{$\infty$} & 7.807 & 7.805 & 8.076 & \multicolumn{1}{c}{$\infty$} & 7.857 & \textbf{7.790} \\
					\bottomrule
				\end{tabular}
			\end{center}
		\end{table}

 
\newpage
\section*{Bibliography} 
\bibliography{bibliography}
\bibliographystyle{abbrvnat}